\documentclass[12pt,a4paper,oneside,onecolumn]{article}

\usepackage{import}

\usepackage{thumbpdf,lmodern}
\usepackage[normalem]{ulem}

\usepackage{framed}
\usepackage{enumerate}
\usepackage{bbm}
\usepackage{amsmath}

\usepackage{tikz}
\usepackage{graphicx}
\usetikzlibrary{positioning}

\usepackage{microtype}
\usepackage{graphicx}

\usepackage{xspace}

\usepackage{amsmath,amssymb,amsfonts}
\usepackage{etoolbox}

\usepackage[thmmarks, amsmath, thref]{ntheorem}
\usepackage{bbm}
\usepackage{bm}

\usepackage[normalem]{ulem} 

\usepackage{times}
\usepackage{epsfig}
\usepackage{graphicx}
\usepackage{xcolor}
\usepackage{subcaption}

\usepackage{diagbox}

\usepackage{lipsum}  

\usepackage{threeparttable}
\usepackage[singlelinecheck=true,justification=centering]{caption}
\usepackage{makecell}
\usepackage{multirow}
\usepackage{color,soul}
\usepackage{xcolor}  

\definecolor{dimgray}{rgb}{0.35, 0.35, 0.35}

\usepackage[linesnumbered,algoruled,noend,noline]{algorithm2e}

\usepackage{libertine}
\usepackage{courier}
\usepackage[utf8x]{inputenc}
\usepackage{multirow}
\usepackage{subcaption}
\usepackage{booktabs}
\usepackage{array}
\usepackage{pdfpages}

\usepackage[longnamesfirst,square,numbers,comma,sort&compress]{natbib}
\usepackage{xcolor}
\usepackage{titlesec}
\usepackage{float}
\usepackage{graphicx}
\usepackage{amsmath}
\usepackage{mathpazo}
\usepackage{datetime}
\usepackage{verbatim}

\RequirePackage{fancyvrb}
\usepackage{hyphenat}

\usepackage[nottoc]{tocbibind}

\usepackage[parfill]{parskip}

\makeatletter
\DeclareRobustCommand\onedot{\futurelet\@let@token\@onedot}
\def\@onedot{\ifx\@let@token.\else.\null\fi\xspace}

\def\eg{\emph{e.g}\onedot} 
\def\ie{\emph{i.e}\onedot} 
 
\def\etc{\emph{etc}\onedot}

\def\iid{\emph{i.i.d}\onedot}

\makeatother

\newcommand{\ZZ}{\ensuremath{\mathbb{Z}}}
\newcommand{\RR}{\ensuremath{\mathbb{R}}}

\newcommand{\Rd}{\ensuremath{\RR^d}}

\newcommand{\Zplus}{\ensuremath{\ZZ^+}}

\newcommand{\set}[1]{\ensuremath{{\{#1\}}}}

\definecolor{orcgreen}{RGB}{100,150,100}

\definecolor{brown}{RGB}{165,42,42}

\bmdefine\ba{a}
\bmdefine\bb{b}
\bmdefine\bc{c}
\bmdefine\bd{d}
\bmdefine\be{e}
\bmdefine\boldf{f}
\bmdefine\bg{g}
\bmdefine\bh{h}
\bmdefine\bi{i}
\bmdefine\bj{j}
\bmdefine\bk{k}
\bmdefine\bl{l}
\bmdefine\bm{m}
\bmdefine\bn{n}
\bmdefine\bo{o}
\bmdefine\bp{p}
\bmdefine\bq{q}
\bmdefine\br{r}
\bmdefine\bs{s}
\bmdefine\bt{t}
\bmdefine\bu{u}
\bmdefine\bv{v}
\bmdefine\bw{w}
\bmdefine\bx{x}
\bmdefine\by{y}
\bmdefine\bz{z}
\bmdefine\bA{A}
\bmdefine\bB{B}
\bmdefine\bC{C}
\bmdefine\bD{D}
\bmdefine\bE{E}
\bmdefine\bF{F}
\bmdefine\bG{G}
\bmdefine\bH{H}
\bmdefine\bI{I}
\bmdefine\bJ{J}
\bmdefine\bK{K}
\bmdefine\bL{L}
\bmdefine\bM{M}
\bmdefine\bN{N}
\bmdefine\bO{O}
\bmdefine\bP{P}
\bmdefine\bQ{Q}
\bmdefine\bR{R}
\bmdefine\bS{S}
\bmdefine\bT{T}
\bmdefine\bU{U}
\bmdefine\bV{V}
\bmdefine\bW{W}
\bmdefine\bX{X}
\bmdefine\bY{Y}
\bmdefine\bZ{Z}

\bmdefine\balpha{\alpha}
\bmdefine\bbeta{\beta}
\bmdefine\bgamma{\gamma}

\bmdefine\bdelta{\delta}
\bmdefine\btheta{\theta}
\bmdefine\blambda{\lambda}
\bmdefine\bphi{\phi}
\bmdefine\bxi{\xi}
\bmdefine\bzeta{\zeta}
\bmdefine\boldeta{\eta}
\bmdefine\bpi{\pi}

\bmdefine\bmu{\mu}
\bmdefine\brho{\rho}
\bmdefine\bomega{\omega}
\bmdefine\bOmega{\Omega}
\bmdefine\bPi{\Pi}

\bmdefine\bvarepsilon{\varepsilon}
\bmdefine\bepsilon{\epsilon}

\bmdefine\bDelta{\Delta}

\bmdefine\bTheta{\Theta}

\bmdefine\bsigma{\sigma}
\bmdefine\bSigma{\Sigma}
\bmdefine\bPsi{\Psi}
\bmdefine\bLambda{\Lambda}

\bmdefine\bzero{0}
\bmdefine\bone{1}
\bmdefine\binfty{\infty}

\newcommand{\Ncal}{\mathcal{N}}

\newcommand{\Wcal}{\mathcal{W}}

\newcommand{\class}[1]{`\code{#1}'}
\newcommand{\fct}[1]{\code{#1()}}

\newcommand{\appropto}{\mathrel{\vcenter{
  \offinterlineskip\halign{\hfil$##$\cr
    \propto\cr\noalign{\kern2pt}\sim\cr\noalign{\kern-2pt}}}}}

\usepackage{hyperref}
\hypersetup{
  colorlinks   = true, 
  urlcolor     = blue, 
  linkcolor    = blue, 
  citecolor   = blue 
}

\DefineVerbatimEnvironment{Code}{Verbatim}{}
\DefineVerbatimEnvironment{CodeInput}{Verbatim}{fontshape=sl}
\DefineVerbatimEnvironment{CodeOutput}{Verbatim}{}
\newenvironment{CodeChunk}{}{}

\DeclareRobustCommand\code[1]{%
  \ifmmode
    \expandafter\texttt
  \else
    \expandafter\textnhtt
  \fi{#1}%
}
\let\proglang=\textsf
\newcommand{\pkg}[1]{{\fontseries{m}\fontseries{b}\selectfont #1}}

\makeatletter
\def\NAT@spacechar{~}
\makeatother

\newcommand{\Plaintitle}[1]{\def\@Plaintitle{#1}}
\newcommand{\Shorttitle}[1]{\def\@Shorttitle{#1}}
\newcommand{\Plainauthor}[1]{\def\@Plainauthor{#1}}
\Plainauthor{\@author}

\date{}

\newtheorem{Example}{Example}

\begin{document}

\title{CPU- and GPU-based Distributed Sampling in
Dirichlet Process Mixtures for Large-scale Analysis}

\author{
  \begin{tabular}{c}
    \shortstack{Or Dinari*$^1$ \vspace{-0.0cm} \\ 
    \href{mailto:dinari@post.bgu.ac.il}{dinari@post.bgu.ac.il}} \\ \\
    \shortstack{John W. Fisher III$^2$ \vspace{-0.0cm} \\
    \href{mailto:fisher@csail.mit.edu}{fisher@csail.mit.edu }}
  \end{tabular}
  \and
    \begin{tabular}{c} 
    \shortstack{Raz Zamir*$^1$ \vspace{-0.0cm} \\
    \href{mailto:razzam@post.bgu.ac.il}{razzam@post.bgu.ac.il}} \\ \\
    \shortstack{Oren Freifeld$^1$  \vspace{-0.0cm}\\
    \href{mailto:orenfr@cs.bgu.ac.il }{orenfr@cs.bgu.ac.il}}
    \end{tabular}
}

\date{%
    $^1$Computer Science, Ben-Gurion University, Beer-Sheva, Israel\\%
    $^2$MIT CSAIL, Cambridge MA, USA%
}

\maketitle

\begin{abstract}
In the realm of unsupervised learning, Bayesian nonparametric mixture models, exemplified by the Dirichlet Process Mixture Model (DPMM), provide a principled approach for adapting the complexity of the model to the data. Such models are particularly useful in clustering tasks where the number of clusters is unknown. Despite their potential and mathematical elegance, however, DPMMs have yet to become a mainstream tool widely adopted by practitioners. This is arguably due to a misconception that these models scale poorly as well as the lack of
high-performance (and user-friendly) software tools that can handle large datasets efficiently. 
In this paper we bridge this practical gap by proposing a new, easy-to-use, statistical software package for scalable DPMM inference. More concretely, we provide efficient and easily-modifiable implementations for high-performance distributed sampling-based inference in DPMMs where the user is free to choose between either 
a multiple-machine, multiple-core, CPU implementation (written in \proglang{Julia}) and a multiple-stream GPU implementation (written in \proglang{CUDA/C++}). Both the CPU and GPU  implementations
come with a common (and optional) python wrapper, providing the user with a single point of entry with the same interface. 
On the algorithmic side, our implementations leverage a leading DPMM sampler from~\cite{Chang:NIPS:2013:ParallelSamplerDP}. While Chang and Fisher III's implementation 
(written in \proglang{MATLAB/C++}) used only CPU and was designed for a single multi-core machine, the packages we proposed here distribute the computations efficiently across either multiple multi-core machines or across mutiple GPU streams. 
This leads to speedups, alleviates memory and storage limitations, and lets us fit DPMMs to significantly larger datasets and of higher dimensionality than was possible previously by either~\cite{Chang:NIPS:2013:ParallelSamplerDP} or other DPMM methods. Our open-source code (GPLv2 licensed) is publicly available on \url{github.com}. 
\end{abstract}

\section{Introduction}
\label{chap:intro}

In unsupervised learning, Bayesian Nonparametric (BNP) mixture models, exemplified by the Dirichlet-Process Mixture Model (DPMM), provide a principled approach for Bayesian modeling while adapting the model complexity to the data. This contrasts with finite mixture models whose complexity is determined manually or via model-selection methods. To fix ideas, an important DPMM example is the Dirichlet-Process Gaussian Mixture Model (DPGMM), a Bayesian $\infty$-dimensional extension of the classical Gaussian Mixture Model (GMM). 
Despite their potential, however, and although researchers have used them successfully in numerous applications during the last two decades, DPMMs still do not enjoy wide popularity among practitioners, largely due to computational bottlenecks that exist in current algorithms and/or implementations. 
In particular, one of the missing pieces is the availability of software tools that: 1) can efficiently handle DPMM inference in large datasets;
2) are user-friendly and can also be easily modified.

We argue that in order \emph{for DPMMs to become a practical choice for large-scale data analysis,
implementations of DPMM inference must leverage parallel- and distributed-computing resources}
(in an analogy, consider how advances in GPU computing and GPU software contributed to the success of deep learning). This is because of not only potential speedups but also memory and
storage considerations. For example, this is especially true in distributed mobile robotic sensing applications where multiple autonomous agents working together have limited computational and communication resources. As another motivating example, consider unsupervised data-analysis tasks in large and high-dimensional computer-vision datasets.

In other words, while DPMMs are theoretically ideal for handling unlabeled datasets, current implementations of DPMM inference do not scale well with the size of the dataset and/or the dimensionality.  

This is partly since most existing implementations are serial and they do not harness the power of distributed computing. This does not mean that there do not exist distributed \emph{algorithms} for DPMM inference. \emph{There is, however, a large practical gap between designing such an algorithm and having it implemented efficiently in a way that fully utilizes the available computing resources}. Thus, the very few publicly-available implementations of such distributed algorithms are fairly limited in their capabilities (as well as their expressiveness; \eg, some support only isotropic Gaussians~\citep{ijcai2017-646}, \etc.).  
Our work closes this gap by providing effective and scalable statistical software for typical large-scale inference.

Concretely, we propose two solutions from which the users can choose according to their needs, constraints, and available computing resources. The first proposed solution, implemented purely in CPU, is based on distributing computations across multiple cores as well as multiple machines. 
The second proposed solution, based mostly on GPU, relies on distributing computations across multiple GPU streams in a single machine. 
See \autoref{tab:overview} for more details. 

\begin{table}[t!]
\centering
\begin{tabular}{llp{8cm}}
\hline
Package             & Processor        & Description \\ \hline
\proglang{CUDA/C++}      & GPU                   & The fastest package for high $N$ (number of data points) and  $d$ (data dimensions) on a single machine; supports multiple GPU streams. \\
\proglang{Julia}    & CPU                        & Supports both multiple cores and multiple machines.\\
\proglang{Python}   & Either CPU or GPU                  & Wrapper to the \proglang{CUDA/C++} and \proglang{Julia} packages \\ \hline

\end{tabular}
\caption{\label{tab:overview} Overview of the proposed packages}
\end{table}

More generally (than the topic of DPMM inference), distributed implementations, at least in traditional programming languages used for such implementations, tend to be hard to debug, read, and modify. This clashes with the usual workflow of algorithm development. 
As a remedy, in our recent workshop paper~\citep{Dinari:CCGRID:2019:distributed}, which constitutes a preliminary and partial
version of this paper, we proposed a \proglang{Julia} implementation
for distributed sampling-based DPMM inference. 
Since the publication of~\citep{Dinari:CCGRID:2019:distributed} we have improved the performance of our Julia implementation, have added its GPU counterpart in \proglang{CUDA/C++}, and added an optional \proglang{Python} wrapper for both the CPU and GPU implementations.

To summarize, in this paper we explain how to use, via either \proglang{Julia} or \proglang{CUDA/C++}, an efficient distributed DPMM inference for large-scale analysis. Particularly, based on a leading parallel Markov Chain Monte Carlo (MCMC) inference algorithm~\citep{Chang:NIPS:2013:ParallelSamplerDP} (to be discussed in~\autoref{chap:models})
-- originally  implemented in \proglang{C++} for a single multi-core CPU single machine in a highly-specialized fashion using a shared-memory model -- we provide novel, more scalable, and easier-to-use-or-modify implementations that leverage either the latest Nvidia's asynch memory allocation API for GPU or \proglang{Julia}'s capabilities
for distributing CPU computations efficiently across multiple multi-core machines using a distributed memory model. This leads to speedups, alleviates memory and storage limitations, and lets us infer DPMMs from significantly larger and higher-dimensional datasets than was previously possible  by either~\cite{Chang:NIPS:2013:ParallelSamplerDP} or other DPMM inference methods.
Our Julia and CUDA/C++ implementations are also accompanied by an optional \proglang{Python} wrapper which hides the \proglang{Julia} \& \proglang{CUDA/C++} code from the user and provides a single point of entry that lets the user employ, in the same settings and with the same code, either one of our CPU and GPU packages.

\section{Models and the Inference Algorithm}
\label{chap:models}
\subsection{Preliminaries: Finite Mixture Models and Clustering}
Let $d$ and $K$ be two positive integers and let $\bx$ be a generic point in $\Rd$. 
A $K$-component Finite Mixture Model (FMM), also known as a parametric mixture model, is a probabilistic model whose associated $d$-dimensional probability density function (pdf)
is
\begin{align}
p(\bx;\theta)=\sum\nolimits_{k=1}^K \pi_k f(\bx;\theta_k)
 \label{Eqn:GMMlikelihood}
 \end{align}
where $\theta=(\pi_k,\theta_k)_{k=1}^K$, $(\pi_k)_{k=1}^K$ form a convex
combination (namely, $\sum_{k=1}^K\pi_k=1$ and $\pi_k\ge0$ $\forall k$), and  $f(\bx;\theta_k$) is a $d$-dimensional pdf parameterized by $\theta_k$. The $(f(\cdot;\theta_k))_{k=1}^K$ functions 
are called the mixture \emph{components}
while $(\pi_k)_{k=1}^K$ are called the mixture \emph{proportions} (or \emph{weights}).
\begin{Example}
In the (finite) Gaussian Mixture Model (GMM), $\theta_k=(\bmu_k,\bSigma_k$)
and $ f(\bx;\theta_k)= \Ncal(\bx;\bmu_k,\Sigma_k)$ where the latter is a multivariate Gaussian pdf
evaluated at $\bx$ and parameterized by a mean vector, $\bmu_k\in\Rd$, and
a Symmetric Positive-Definite (SPD) $d\times d$ covariance matrix, $\bSigma_k$. In other words, in this example
\begin{align}
f(\bx;\theta_k)=
\Ncal(\bx;\bmu_k,\bSigma_k)\triangleq 
(2\pi)^{-d/2}(\det \bSigma_k)^{-1/2}\exp(-\tfrac{1}{2} (\bx-\bmu_k)^T \bSigma_k^{-1}(\bx-\bmu_k))\,. 
\end{align}
\end{Example}
The case where $\bx$ takes values in a discrete set is similar, except that each component is a probability mass function (pmf), not a pdf. A popular example is a mixture of categorical distributions or, more generally, multinomial distributions. 

Let $\bX=(\bx_i)_{i=1}^N$ stand for $N$ data points and again let $K>0$ be an integer. 
\emph{Clustering} is the task of partitioning
$\bX$ into $K$ parts, called \emph{clusters} and denoted by $\bC=(C_k)_{k=1}^K$. Let $z_i$ denote the 
latent point-to-cluster assignment of $\bx_i$; \ie, cluster $k$ is defined as $C_k=(\bx_i)_{i:z_i=k}$. 
In the standard formulation of FMM-based clustering, 
the data points are assumed to be independent and identically-distributed (\iid) draws from  
an FMM and one typically tries to maximize the corresponding 
likelihood function, $p(\bX|\theta)$, over  $\theta$.
The most popular approach for (locally-) maximizing that likelihood  is via 
\emph{an} Expectation-Maximization (EM) algorithm~\citep{Dempster:JRSS:1977:EM}.  
As the time-honored Bayesian formulation helps avoiding problems such as over-fitting and enables encoding prior knowledge, the  FMM also has a Bayesian (but still finite)
variant, where $\theta$ is assumed to be random and drawn from a suitable prior~\citep{Gelman:Book:2013:Bayesian} (specifically, the prior over $(\pi_k)_{k=1}^K$ is usually a Dirichlet \emph{distribution}, not to be confused with a Dirichlet \emph{process}). In which case, one targets the \emph{posterior} distribution,
$p(\theta|\bX)$, 
rather than the likelihood; \eg, one can try to maximize $p(\theta|\bX)$ over $\theta$, sample $\theta$ from it, compute posterior expectations, \etc.  

\subsection{The Dirichlet Process Mixture Model (DPMM)}
Below we provide a brief 
and \emph{informal} introduction to the DPMM, focusing only on the essentials required for understanding this manuscript and make it self-contained as possible.
For a comprehensive and formal mathematical treatment, see~\cite{Ghosal:Book:2017:BNP}. For an applied data-analysis perspective, see~\cite{Muller:BOOK:2015:BNP}.  
For a gentle introduction with machine-learning and/or computer-vision readers in mind, see the theses by~\cite{Sudderth:PhD:2006:GraphicalModels} or~\cite{Chang:Thesis:2014:Sampling}.

The DPMM is a BNP extension of the FMM~\citep{Antoniak:AoS:1974:DPMM}.
Loosely speaking, a DPMM entertains the notion of a mixture of infinitely-many \emph{components}. 

The weights, $\bpi=(\pi_k)_{k=1}^\infty$, are drawn from a Griffiths-Engen-McCloskey (GEM) stick-breaking process with a concentration parameter $\alpha>0$~\citep{Pitman:Book:2002:Combinatorial},
while the components, $(\theta_k)_{k=1}^\infty$, are drawn from a prior as in the Bayesian FMM case.
\begin{Example}
The Dirichlet Process GMM (DPGMM) is a GMM with infinitely-many Gaussians.
\end{Example}
Like the FMM, the DPMM is often used for clustering.
With a DPMM, however, the number of \emph{clusters}, $K$, is not assumed to be known;
rather, it is viewed as a latent random variable whose value is inferred with the rest of the model. 

While in an FMM
the number of clusters and the number of components are typically
equal, and are thus both denoted by a single symbol, $K$, with a DPMM
the situation is different: although there are infinitely-many components,
the (latent and random) number of clusters, $K$, is finite (particularly,  $K\le N$).
The inferred value of $K$ depends on $\alpha$ (the higher $\alpha$ is, the more clusters are expected), on the complexity of the data, and the (hyper-parameters of the) prior over the components (that said, when $N$ is large,
the last two factors are usually more influential than $\alpha$). 
\begin{Example}
Recall that the standard prior
for Gaussian components is a Normal Inverse-Wishart (NIW) distribution~\citep{Gelman:Book:2013:Bayesian}. If the NIW's hyper-parameters 
are set to strongly favor small covariances (hence small clusters), then this will implicitly favor a large $K$. Likewise, if the NIW prior strongly favors larger covariances (hence large clusters), then $K$ will tend to be small. In the lack of prior knowledge, however, the NIW prior can be set to be very weak (\ie, high uncertainty), letting the data speak for itself. 
\end{Example}

For simplicity, our text below implicitly assumes that all the random vectors involved have either a pdf or a pmf. One known mathematical construction of the DPMM uses the following distributions: 
\begin{align}
& \bpi|\alpha \sim \mathrm{GEM}(\bpi;\alpha), \label{Eqn:DPLatentVar:pi}\\
& \theta_k |\lambda  \overset{\iid}{\sim} f_\theta(\theta_k;\lambda), \qquad \forall k\in \set{1,2,\ldots}, \\
& z_i|\bpi \overset{\iid}{\sim} \mathrm{Cat}(z_i;\bpi) \qquad \forall i\in \set{1,2,\ldots,N}, \\
& \bx_i|z_i,\theta_{z_i} \sim f_\bx(\bx_i;\theta_{z_i}), \qquad\forall i\in \set{1,2,\ldots,N}\,. \label{Eqn:DPLatentVar:xi}
\end{align}
Here, $f_\theta(\cdot;\lambda)$ is the pdf or pmf (associated with some base measure)  parameterized by
$\lambda$,   
the infinite-length vector $\bpi=(\pi_k)_{k=1}^\infty$ is drawn from a GEM stick-breaking process with a concentration parameter $\alpha>0$ (particularly, $\pi_k>0$ for every $k$ and $\sum_{k=1}^{\infty}\pi_k=1$) 
while $\theta_k$ is drawn from $f_\theta$. 
Each of the $N$ \iid  observations $(\bx_i)_{i=1}^N$ is generated 
by first drawing a label, $z_i\in \Zplus$, from $\bpi$ (\ie, $ \mathrm{Cat}$ is the categorical distribution), and then  $\bx_i$ is drawn from (a pdf or a pmf) $f_\bx$ parameterized by $\theta_{z_i}$. 
Informally,
\begin{align}
 \bx_i \overset{\iid}{\sim} \sum\nolimits_{k=1}^{\infty} \pi_k f_\bx(\bx_i;\theta_k) \, .
\end{align}
Here too, each $f_\bx(\cdot;\theta_k)$ is called a \emph{component} and we make no distinction between a component, $f_\bx(\cdot,\theta_k)$, and its parameter, $\theta_k$. 
The so-called labels $(z_i)_{i=1}^N$ encode
the observation-to-component assignments. 
A cluster is a collection of points sharing a label; \ie, $\bx_i$ is in cluster $k$, denoted by $C_k$,
if and only if $z_i=k$. 
Let (the random variable) $K$
be the number of unique labels: $K=|\set{k:z_i=k \text{ for some } i\in \set{1,\ldots,N}}|$; \ie, $K$ is also the number of clusters and is bounded above by $N$. 
Typically, and as assumed in this manuscript, $f_\theta$ is chosen to be a conjugate prior~\citep{Gelman:Book:2013:Bayesian} to $f_\bx$.
The latent variables here are $K$, $(\theta_k)_{k=1}^\infty$, $\bpi$, and $(z_i)_{k=1}^N$.
For more details (and other constructions), see~\cite{Sudderth:PhD:2006:GraphicalModels}.
\begin{Example}
 In the case of Gaussian components, where $f_\theta(\cdot;\lambda)$
 is an NIW pdf, we have 
 {
        \begin{align}
        &f_\theta(\overbrace{\bmu_k,\bSigma_k}^{\theta_k};\overbrace{\kappa,\bm,\nu,\bPsi}^{\lambda})
        =        
       \mathrm{NIW}(\bmu,\bSigma;\kappa,\bm,\nu,\bPsi)
       \triangleq         
\Ncal(\bmu;\bm,\tfrac{1}{\kappa}\bSigma)
\Wcal^{-1}(\bSigma;\nu,\bPsi)
        \end{align} 
 }
where $\Wcal^{-1}(\bSigma;\nu,\bPsi)$ is 
the Inverse-Wishart distribution (over $d\times d$ SPD matrices),
the hyperparamers are 
\begin{align}
 \lambda=(\bm,\bPsi,\kappa,\nu)
\end{align}
where $\bm\in\Rd$, the $d\times d$ matrix $\bPsi$ is SPD, and the two real numbers $\kappa$ and $\nu$ satisfy $\kappa>0$ and $\nu>d-1$ (do not confuse $\kappa$ (``kappa'') with $k$, the index of the component).  
\end{Example}

\subsection{Inference via Chang and Fisher III's DPMM Sampler}
We now briefly review a DPMM sampler proposed by~\cite{Chang:NIPS:2013:ParallelSamplerDP}.
That sampler consists of a restricted Gibbs sampler~\citep{Robert:Book:2013:Monte}
and a split/merge framework~\citep{jain2004split} which together form an ergodic Markov chain.
The operations in each step of that sampler are highly parallelizable. 
Importantly, the splits and merges let the sampler make \emph{large moves}
along the (posterior) probability surface as in such operations multiple labels change
their label \emph{simultaneously} to the same different label. This is unlike what happens, \eg, in methods that must change each label separately from the others. We now describe the essential details. 

\textbf{The augmented space.}
The latent variables, 
$(\theta_k)_{k=1}^\infty$, $\bpi$, and $(z_i)_{k=1}^N$, 
are augmented with auxiliary variables. 
For each component $\theta_k$ two subcomponents (conceptually thought of as $l$=``left'' and $r$=``right) are added, $\bar{\theta}_{k,l},\bar{\theta}_{k,r}$, with subcomponent weights $\bar{\bpi}_k=(\bar{\pi}_{k,l},\bar{\pi}_{k,r})$. 
Implicitly, this means that every cluster $C_k$ is  augmented with two subclusters, $\bar{C}_{k,l}$ and $\bar{C}_{k,r}$. 
For each cluster label $z_i$, an additional \emph{subcluster label}, $\bar{z}_i\in\set{l,r}$, is added; \ie, subcluster $\bar{C}_{k,l}\subset C_{k}$ consists of all the points in $C_k$ whose subcluster label is $l$ (the other subcluster, $\bar{C}_{k,r}$, is defined similarly). 

\textbf{The restricted Gibbs sampler.}
This restricted sampler is not allowed to change (the current estimate of) $K$; 
rather, it can change only 
the parameters of the existing clusters and subclusters, and when sampling the labels,
it can assign an observation only to an existing cluster.
For each instantiated component $k$, changing \begin{align}\theta_k, \bar{\theta}_{k,l}, \text{ and }\bar{\theta}_{k,r}\end{align}
is done using 
\begin{align}
 p(\theta_k|C_k;\lambda), 
 p(\bar{\theta}_{k,l}|\bar{C}_{k,l};\lambda), \text{ and }
 p(\bar{\theta}_{k,r}|\bar{C}_{k,r};\lambda)
 \,,\end{align} respectively, where the latter three are the conditional distributions of the cluster or subcluster parameters given the cluster or subclusters (and the prior hyperparamers, $\lambda$) . 
In~\autoref{chap:Detailed Implementation Design}, we will dive 
deeper into the details of the restricted Gibbs sampler.

\textbf{The split/merge framework.}
Splits and merges allow the sampler to change $K$ 
using the Metropolis-Hastings framework~\citep{Hastings:1970:MC}. 
Particularly, the auxiliary variables are used to propose splitting an existing 
cluster or merging two exiting ones. 
When a split is accepted, each of the newly-born
clusters is augmented with two new subclusters.
The Hastings ratio of a split is~\citep{Chang:NIPS:2013:ParallelSamplerDP}:
\begin{align}
 \mathrm{H}_{\text{split}}=\frac{\alpha 
 \Gamma(N_{k,l})f_\bx(\bar{C}_{k,l};\lambda)
 \Gamma(N_{k,r})f_\bx(\bar{C}_{{k},2};\lambda)}{\Gamma(N_{k})f_\bx(C_k;\lambda)}
 \label{Eqn:HastingsRatioSplit}
\end{align}
where $\Gamma$ is the Gamma function,
$N_k$, $N_{k,l}$ and $N_{k,r}$ are the numbers of points in
$C_k$, $\bar{C}_{k,l}$ and $\bar{C}_{k,r}$, respectively,
while \begin{align}f_\bx(C_k;\lambda), f_\bx(\bar{C}_{k,l};\lambda), \text{ and } f_\bx(\bar{C}_{k,r};\lambda)\end{align} represent the \emph{marginal}
likelihood of $C_k$, $\bar{C}_{k,l}$ and $\bar{C}_{k,r}$ respectively. Concrete expressions for the marginal likelihood, in the case of Gaussian or Multinomial components
(the component types considered in our experiments) appear in~\cite{Chang:Thesis:2014:Sampling}.

Finally, a \emph{merge} proposal is based on taking two existing clusters
and proposing merging them into one. The corresponding Hastings ratio is
 $\mathrm{H}_{\text{merge}} =1 / \mathrm{H}_{\text{split}}$ where $\bar{C}_{k,l}$ and $\bar{C}_{k,r}$
  are replaced with the two clusters, and $C_k$ is replaced with the result of the merge.
For the derivation behind  $\mathrm{H}_{\text{split}}$ and  $\mathrm{H}_{\text{merge}}$, see~\cite{Chang:NIPS:2013:ParallelSamplerDP}.

\begin{figure}
  \begin{subfigure}[b]{0.5\columnwidth}
    \includegraphics[width=\linewidth]{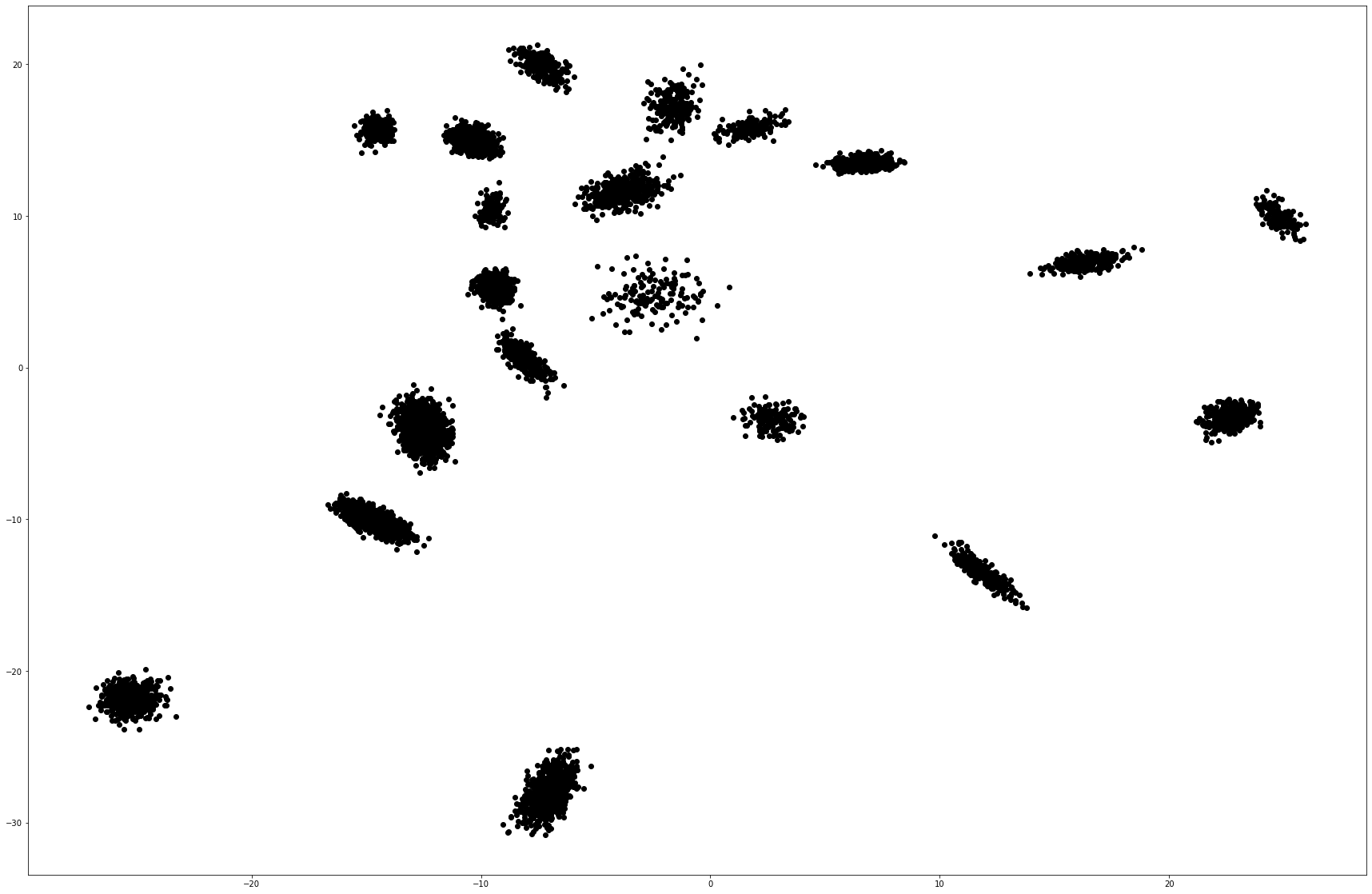}
    \caption{Unlabeled data}
    \label{fig:sub1}
  \end{subfigure}
  \hfill 
  \begin{subfigure}[b]{0.5\columnwidth}
    \includegraphics[width=\linewidth]{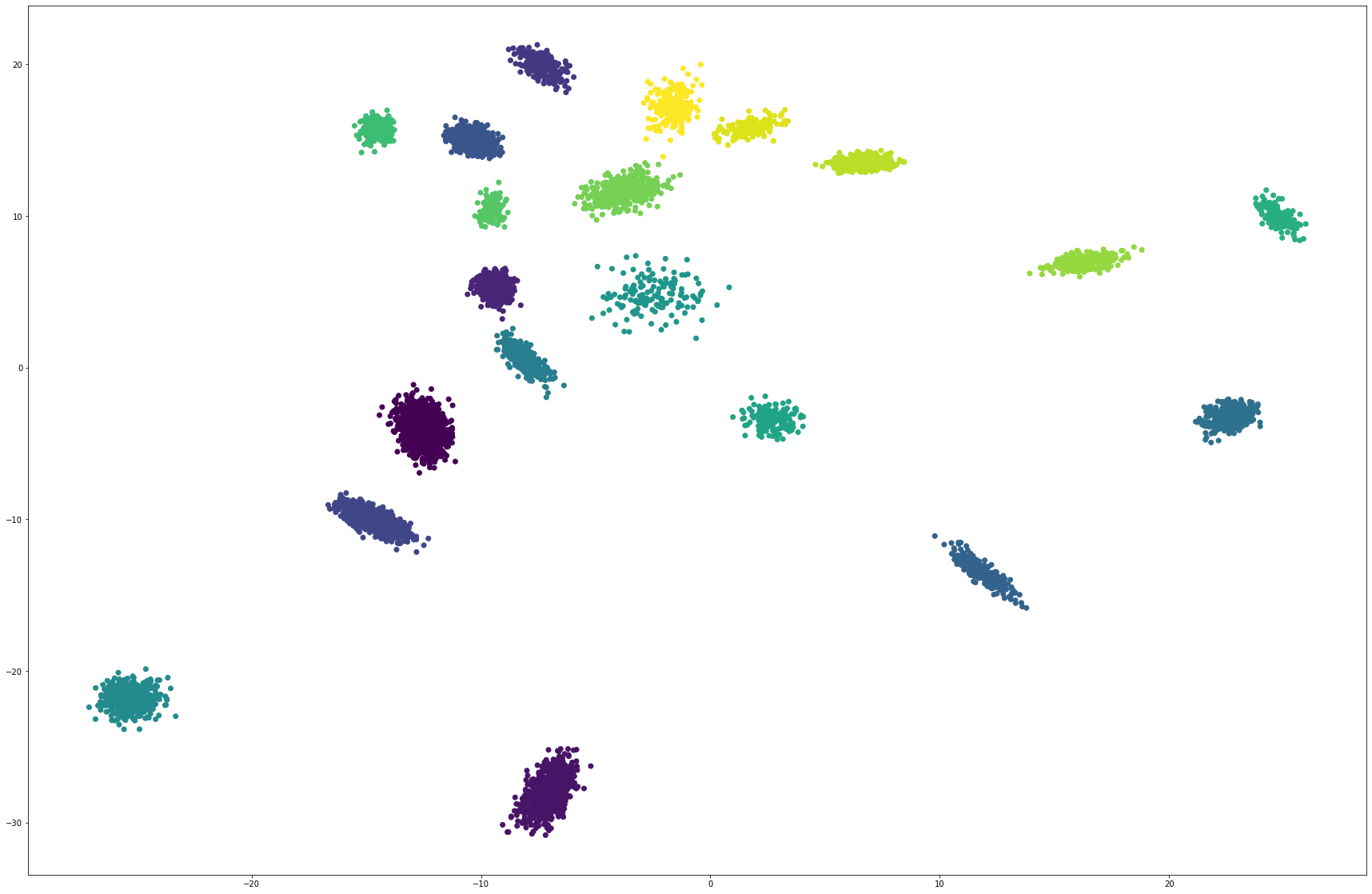}
  \caption{DPMMSubClusters Detection}
  \label{fig:sub2}
  \end{subfigure}
\caption{20 clusters detected by DPMMSubClusters}
\label{fig:clustering20}
\end{figure}

\begin{figure}
  \begin{subfigure}[b]{0.55\columnwidth}
    \includegraphics[width=\linewidth]{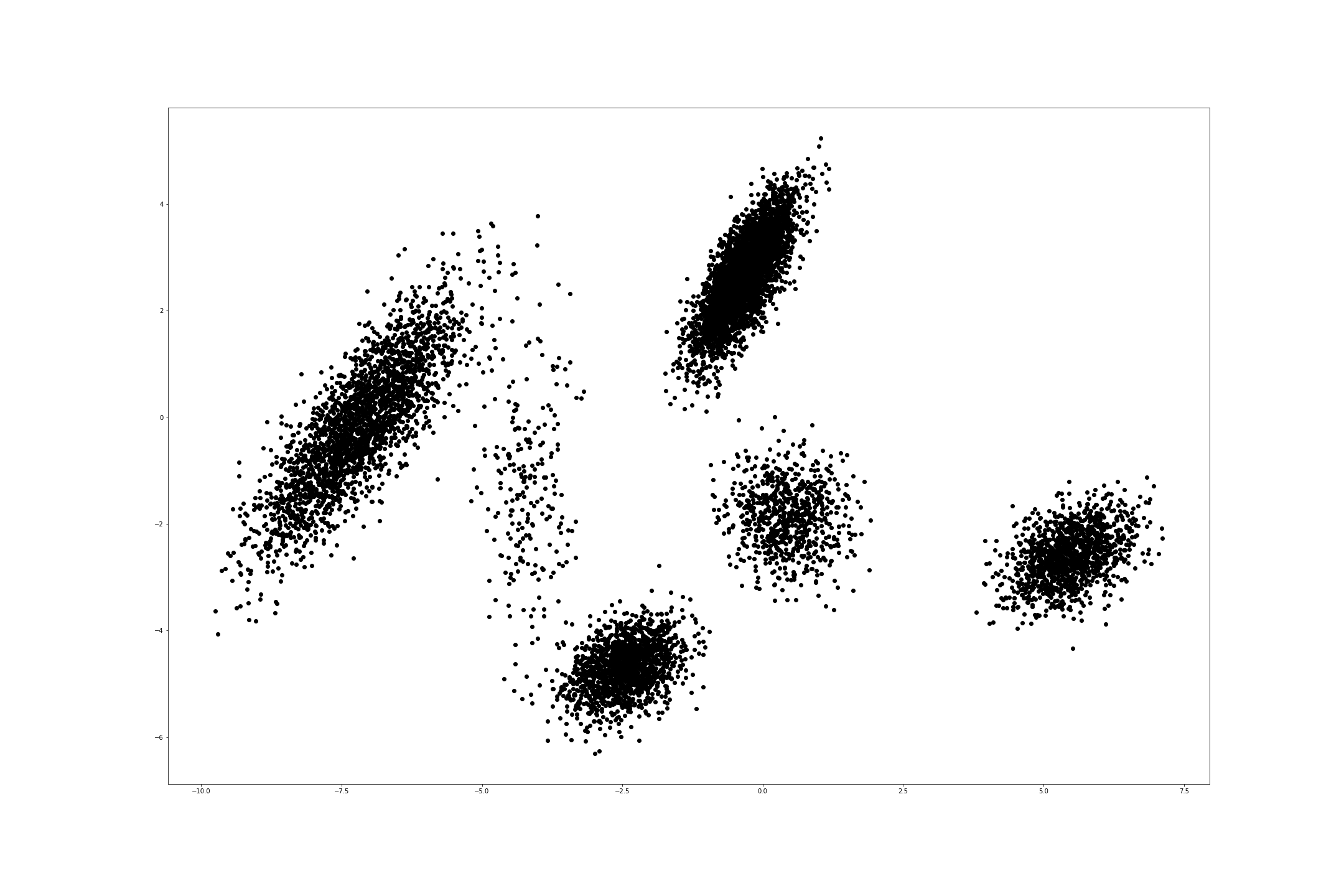}
    \caption{Unlabeled data}
    \label{fig:sub3}
  \end{subfigure}
  \hfill 
  \begin{subfigure}[b]{0.55\columnwidth}
    \includegraphics[width=\linewidth]{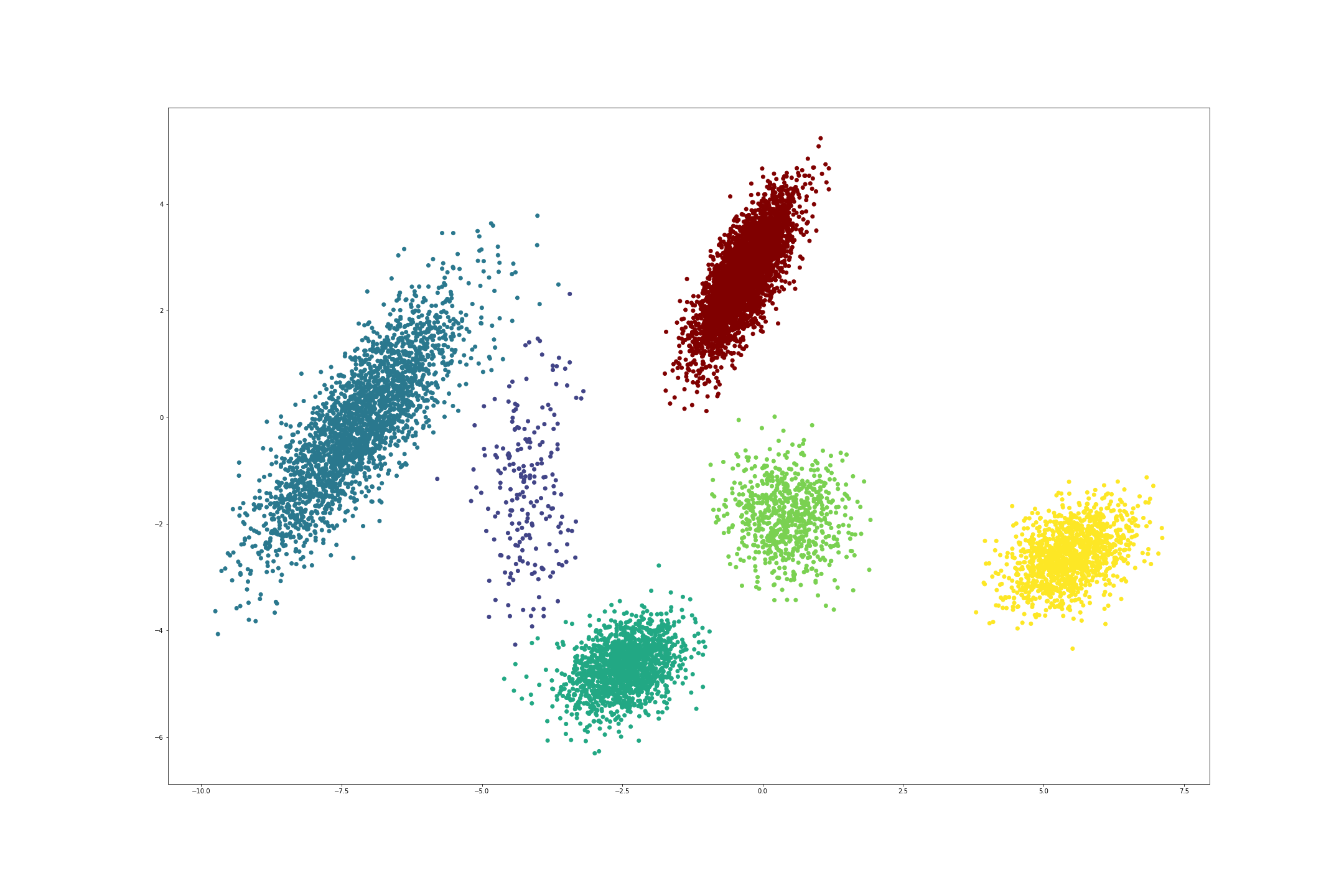}
  \caption{DPMMSubClusters Detection}
  \label{fig:sub4}
  \end{subfigure}
\caption{6 clusters detected by DPMMSubClusters}
\label{fig:clustering6}
\end{figure}

Importantly, and as any successful DPMM inference method, Chang and Fisher III's sampler can detect different numbers of clusters according to the
complexity of the dataset. For example, in~\autoref{fig:clustering20} we demonstrate unlabeled data with 20 clusters while in \autoref{fig:clustering6} we show data consisting of 6 clusters. 
Using our implementation (the topic of the next section), the sampler
correctly detected the different numbers of clusters
in each dataset, while using the same code and the same hyperparamters. 

\section{Software}

The proposed software supports three different software languages (\proglang{Julia}, \proglang{CUDA/C++}, \proglang{Python}), two operating systems (Windows \& Linux) as well as multiple interface options.  In terms of performance, the proposed software is faster than other publicly-available packages that exist today and that we were able to test. 
Particularly, as long as the product of $N$ (number of data points) times $d$ (the dimension of the data) is not too low, then our GPU version is consistently at least a few times faster than any other implementation we
are aware of.

\subsection{Installation}
Our software (licensed under GPLv2) is available on GitHub. The URLs for our packages appear in~\autoref{tab:github}. In order to compile the \proglang{CUDA/C++} package the user will need \proglang{C++14} (or higher) and \proglang{CUDA} 11.2 (or higher). 
For visualization purposes (\ie, plotting points in 2D), the \proglang{CUDA/C++} windows version
also requires \pkg{OpenCV}. However, this visualization is used only for debugging purposes so the installation
of \pkg{OpenCV} is not mandatory for using our implementation. For the CPU package the user will need \proglang{Julia} 1.5 (or higher). For \proglang{Python} we used 3.8. 
The installation steps below were tested successfully on Windows 10 (Visual Studio 2019), Ubuntu 18.04, and Ubuntu 21.04.

\begin{enumerate}[(1)]
    \item Install \proglang{CUDA} version 11.2 (or higher) from \\ \url{https://developer.nvidia.com/CUDA-downloads}
    \item git clone \url{https://github.com/BGU-CS-VIL/DPMMSubClusters_GPU}
    \item Install Julia from: \url{https://julialang.org/downloads/platform}
    \item Add our DPMMSubCluster package from within a Julia terminal via Julia package manager:  \\  \code{] add DPMMSubClusters}
    \item Add our dpmmpython package in python: \\
    pip install dpmmpython
    \item Add Environment Variables:
            \begin{itemize}\item
    On Linux:
        \begin{enumerate}[(a)]
            \item Add "CUDA\_VERSION" with the value of the version of your \proglang{CUDA} installation (\eg, 11.5).
            \item Add to the "PATH" environment variable the path to the \proglang{Julia} executable (\eg, in .bashrc add: export PATH =\$PATH:\$HOME/julia/julia-1.6.0/bin).
            \item Make sure that CUDA\_PATH exist. If it is missing add it with a path to CUDA (\eg, export CUDA\_PATH=/usr/local/cuda-11.6/).
            \item Make sure that the relevant CUDA paths are included in \$PATH and \$LD\_LIBRARY\_PATH (\eg, export PATH=/usr/local/cuda-11.6/bin:\$PATH, export LD\_LIBRARY\_PATH=/usr/local/cuda-11.6/lib64:\$LD\_LIBRARY\_PATH).

        \end{enumerate}
\item        On Windows:
        \begin{enumerate}[(a)]
            \item Add "CUDA\_VERSION" with the value of the version of your \proglang{CUDA} installation (\eg, 11.5).
            \item Add to the "PATH" environment variable the path to the \proglang{Julia} executable\\ (\eg, C:\verb+\+Users\verb+\+<USER>\verb+\+AppData\verb+\+Local\verb+\+Programs\verb+\+Julia\verb+\+Julia-1.6.0\verb+\+bin).
            \item Make sure that CUDA\_PATH exists. If it is missing add it with a path to CUDA (\eg, C:\verb+\+Program Files\verb+\+NVIDIA GPU Computing Toolkit\verb+\+CUDA\verb+\+v11.6).
        \end{enumerate}
        \end{itemize}
    \item Install cmake if necessary. 
    \item Install PyJulia  from within a \proglang{Python} terminal: \\ \code{import julia;julia.install();}
    \item For Windows only (optional, used on for debugging purposes): Install \pkg{OpenCV}
        \begin{enumerate}[(a)]
        \item run Git Bash
        \item cd <YOUR\_PATH\_TO\_DPMMSubClusters\_GPU>/DPMMSubClusters
        \item ./installOCV.sh
    \end{enumerate}
\end{enumerate}

\begin{table}[t!]
\centering
\begin{tabular}{ll}
\hline
Package             & GitHub \\ \hline
\proglang{CUDA/C++} & \url{https://github.com/BGU-CS-VIL/DPMMSubClusters_GPU}\\
\proglang{Julia}    & \url{https://github.com/BGU-CS-VIL/DPMMSubClusters.jl}\\
\proglang{Python}   & \url{https://github.com/BGU-CS-VIL/dpmmpython}\\ \hline
\end{tabular}
\caption{\label{tab:github} GitHub URL for each package.}
\end{table}

\subsection{Building}
For Windows for the \proglang{CUDA/C++} package both of the build options below are viable. For Linux use Option 2.\\
Option 1: DPMMSubClusters.sln - Solution file for Visual Studio 2019\\
Option 2: CMakeLists.txt
\begin{enumerate}[(1)]
    \item Run in the terminal:
\begin{CodeChunk}
\begin{CodeInput}
cd <YOUR_PATH_TO_DPMMSubClusters_GPU>/DPMMSubClusters
mkdir build
cd build
cmake -S ../
\end{CodeInput}
\end{CodeChunk}
    \item Build:\\
    Windows:
\begin{CodeChunk}
\begin{CodeInput}
cmake --build . --config Release --target ALL_BUILD
\end{CodeInput}
\end{CodeChunk}
    Linux:
\begin{CodeChunk}
\begin{CodeInput}
cmake --build . --config Release --target all
\end{CodeInput}
\end{CodeChunk}
\end{enumerate}
    
\subsection{Post Build}
Add Environment Variable:
\begin{itemize}
    \item On Linux:\\
        Add "DPMM\_GPU\_FULL\_PATH\_TO\_PACKAGE\_IN\_LINUX" with the value of the path to the binary of the package DPMMSubClusters\_GPU.\\ The path is: \textless YOUR\_PATH\_TO\_DPMMSubClusters\_GPU\textgreater/\\DPMMSubClusters/DPMMSubClusters.
    \item On Windows:\\
        Add "DPMM\_GPU\_FULL\_PATH\_TO\_PACKAGE\_IN\_WINDOWS" with the value of the path to the exe of the package DPMMSubClusters\_GPU.\\
        The path is: \textless YOUR\_PATH\_TO\_DPMMSubClusters\_GPU\textgreater\verb+\+DPMMSubClusters\verb+\+\\build\verb+\+Release\verb+\+DPMMSubClusters.exe.
\end{itemize}

\subsection{Executing}
Below are listed several options to run DPMM inference using our software. Usually the best performance will be achieved by running the \proglang{CUDA/C++} version which uses the GPU (when available). 
\subsubsection{From Julia code - CPU}
The following sample code generates a synthetic GMM dataset with $N=10^5$ points, dimension $d=2$ and $K=10$ clusters,
and then fits a DPMM to the data (without knowing $K$ or the other parameters of the GMM) using the sampler. 

\begin{Code}
using DPMMSubClusters

x,labels,clusters = 
    DPMMSubClusters.generate_gaussian_data(10^5, 2, 10, 100.0)
hyper_params = 
    DPMMSubClusters.niw_hyperparams(Float32(1.0), 
                                    zeros(Float32,2),
                                    Float32(5), 
                                    Matrix{Float32}(I, 2, 2)*1)
DPMMSubClusters.dp_parallel(x, hyper_params, Float32(100000.0), 
                            100, 1, nothing, true, false, 15,
                            labels)
\end{Code}

\subsubsection{From a C++ code -- GPU}
\fct{dp\_parallel} in \class{dp\_parallel\_sampling\_class} is the function that should be called in order to run the program. The sample code below will first generate a synthetic random dataset and will then run the sampler to fit a DPMM to it. 
\begin{Code}
srand(12345);
data_generators data_generators;
MatrixXd  x;
std::shared_ptr<LabelsType> labels =
    std::make_shared<LabelsType>();
double** tmean;
double** tcov;
int N = (int)pow(10, 5);
int D = 2;
int numClusters = 2;
int numIters = 100;

data_generators.generate_gaussian_data(N, D, numClusters, 100.0,
                                       x, labels, tmean, tcov);
std::shared_ptr<hyperparams> hyper_params = 
    std::make_shared<niw_hyperparams>(1.0, VectorXd::Zero(D), 5, 
                                      MatrixXd::Identity(D, D));
dp_parallel_sampling_class dps(N, x, 0, prior_type::Gaussian);
ModelInfo dp = dps.dp_parallel(hyper_params, N, numIters, 1, true,
                               false, false, 15, labels);
\end{Code}

\subsubsection{From the command line -- GPU}
Running the \proglang{CUDA/C++} program can be done from the command line (in both Linux and Windows). There are a few parameters that can be used to run the program. In order to set the hyperparams the \code{params\_path} parameter can be used. The value of the parameter should be a path for a Json file which includes the hyperparams (\ie alpha or hyper\_params for the prior) . In order to use this parameter follow this syntax:
\begin{CodeChunk}
\begin{CodeInput}
--params_path=<PATH_TO_JSON_FILE_WITH_MODEL_PARAMS>
\end{CodeInput}
\end{CodeChunk}
There are few more parameters like \code{model\_path} for the path to a npy file which include the model and \code{result\_path} for the path of the output results.

Our code has support for both Gaussian and Multinomial distributions. It can be easily adapted to other component distributions, \eg, Poisson, as long as they belong to an exponential family. The default distribution is Gaussian. To specify a distribution other than a Gaussian, use the \code{prior\_type} parameter. For example:
\begin{CodeChunk}
\begin{CodeInput}
--prior_type="Multinomial"
\end{CodeInput}
\end{CodeChunk}
The Json file containing the model parameters can contain many parameters that can be controlled by the user.
A few examples are: alpha, prior, number of iterations, burn\_out and kernel. The full list of parameters can be seen in the function \fct{init} in \class{global\_params}. The result file is a Json file which by default contains the predicted labels, the weights, the Normalized Mutual Information (NMI) score and the running time per iteration. A few other parameters can be added to the result file. Samples for these additional parameters are commented out in the main.cpp file.

\subsubsection{From Python code -- CPU and GPU}
The dpmmwrapper.py file contains an example for how to run either the GPU or CPU packages from \proglang{Python} (look for the code in the function \code{main}). In the following example we are generating a GMM synthetic dataset for $10^5$ point, 2 dimensions and 10 clusters. We are running it on the GPU.
\begin{Code}

from julia.api import Julia
jl = Julia(compiled_modules=False)
from dpmmpython.dpmmwrapper import DPMMPython
from dpmmpython.priors import niw

data, gt = DPMMPython.generate_gaussian_data(sample_count=10000,
                                             dim=2, k=10, 
                                             var=100.0)
prior = niw(kappa = 1, mu = np.zeros(2), nu = 3, psi = np.eye(2))
labels, clusters, results = DPMMPython.fit(data = data,
                                    alpha = 100,
                                    prior = prior, verbose = True,
                                    gt = gt,
                                    gpu = True)
\end{Code}
Here, the variable ``results'' depends on the backend: for the \proglang{CUDA/C++} backend it will be `null`,  while for the \proglang{Julia} backbone ``results'' will contain other information (see the documentation of our \proglang{Julia} `fit' function). The ``gt'' variable stands for the ground-truth labels. 

\subsubsection{From a Jupiter notebook -- CPU and GPU}
The notebook dpmmwrapper.ipynb can be executed to test different packages including our GPU and CPU packages on
multiple datasets.

\subsection{Test}
Our \proglang{CUDA/C++} package was built with the Test Development Driven (TDD) methodology. The Windows version comes with 53 unit tests which cover more than 60\% of the code. Usually it takes less than 3 minutes to run all tests. Those tests are for both Gaussian and Multinomial distributions, and include the different CUDA  kernels  (for matrix multiplication) mentioned in~\autoref{chap:Detailed Implementation Design} below. We used the GoogleTest framework to write those tests and they were validated with  a Visual Studio 2019 test engine.

\section{The Proposed Implementation}
\label{chap:Detailed Implementation Design}
\subsection{Design and Implementation}
In our implementation(s) after we initialize all the required objects, we start running the iterations of the sampler. For each iteration in \fct{group\_step} we execute the algorithm which is described below in detail. For concreteness, below we will refer to the \proglang{CUDA/C++} code during the algorithm's description and not to the \proglang{Julia} code. However, the structure of the code in both packages is similar enough to be followed as long as the reader is familiar with any of those languages.

We use the same notation as in~\cite{Chang:NIPS:2013:ParallelSamplerDP} where $N$ is the number of data points and $K$ is the (current estimate of the) number of clusters.
\begin{itemize}
\item For each iteration of the restricted Gibbs sampling:
\begin{enumerate}[(a)]
\item Sample cluster weights $\pi_1,\pi_2,\dots,\pi_K$:
\begin{align} \label{eq:sample_weights}
(\pi_1,\dots,\pi_K,\tilde \pi_{K+1}) \sim \text{Dir}(N_1, \dots, N_K, \alpha).
\end{align}
In our code we built a vector \code{points\_count} that holds the number of points for each cluster and apply a Dirichlet-distribution sampling at once for all clusters:   
\begin{Code}
dirichlet_distribution<std::mt19937> d(points_count);
std::vector<double> dirichlet = d(*globalParams->gen);
\end{Code}
\item 
Sample sub-cluster weights from a 2D Dirichlet distribution:  
\begin{align} \label{eq:sample_sub_weights}
(\bar \pi_{kl}, \bar \pi_{kr}) \sim \text{Dir}(N_{kl} + \alpha/2, N_{kr} + \alpha/2), \quad \forall k \in \{1,\dots,K\}.
\end{align}
In order to propose meaningful splits that are likely to be accepted, the algorithm uses auxiliary variables such that each cluster consists of 2 sub-clusters (conceptually thought of as ``left'' and ``right''). $\bar{\pi}_k = \{\bar{\pi}_{kl},\bar{\pi}_{kr}\}$ denote the weights of the sub-clusters of cluster $k$. The code is implemented in \fct{sample\_cluster\_params} in \class{shared\_actions}.

\item Sample cluster parameters: 
\begin{align} \label{eq:sample_params}
\theta_k \appropto f_{\bx}(\bx_{\mathcal{I}_k};\theta_k) f_\theta(\theta_k;\lambda), \quad \forall k: \in \{1,\dots,K\}.
\end{align}
Here, $f_{\bx}(X;\theta_k)$ is the likelihood of a set of data points under the parameter $\theta_k$, $f_\theta(\theta;\lambda)$ is the likelihood of the parameter $\theta$ under the prior $f_\theta(\cdot;\lambda)$, $\appropto$ denotes sampling proportional to the right-hand side of the equation, and $\mathcal{I}_k=\{i:z_i=k\}$ is the set of indices of that points labels as belonging to cluster $k$.
Each distribution (\eg, a Gaussian or a Multinomial) has its own \fct{sample\_distribution} function which calculates the cluster's parameters. The prior classes for the Gaussian and Multinomial distributions are \class{niw} and the \class{multinomial\_prior} respectively which inherit from the \class{prior} class. The likelihood is calculated in each class within the function \fct{log\_marginal\_likelihood}. In order to extend and support to more distributions new classes which inherit from \class{prior} may be added.

\item Sample sub-cluster parameters: 
\begin{align} \label{eq:sample_sub_params}
\bar\theta_{kh} \appropto f_{\bx}(\bx_{\mathcal{I}_{kh}};\bar\theta_{kh}) f_\theta(\bar\theta_{kh};\lambda), \quad \forall k: \in \{1,\dots,K\} \quad , \quad h \in \{l,r\}.
\end{align}
$\bar \theta_k = \{\bar \theta_{kl}, \bar \theta_{kr}\}$ denotes the parameters of the sub-clusters of cluster $k$.
In the code we used the same functions that we used for the calculation the cluster's parameters.
We maintain the following objects in memory: a \code{std::vector} of \code{local\_cluster} type where each item in that vector holds the cluster and sub-cluster parameters (\eg, $\bar\theta_k$, $\mu_k$, $\Sigma_k$,$\bar\theta_{kl}$, $\bar\theta_{kr}$, $\bar\pi_{kl}$ and $\bar\pi_{kr}$ for the $k^\text{th}$ Gaussian), the point counts (\ie, $N_k$, $N_{kl}, N_{kr}$) and the sufficient statistics as well as the cluster weights (\ie, $\pi_1,\pi_2,\ldots,\pi_k$).

\item Sample cluster assignments for each point: 
\begin{align} \label{eq:sample_labels}
z_i \appropto \sum\nolimits_{k=1}^{K} \pi_k  f_{\bx}(\bx_i;\theta_{k}) \mathbbm{1}(z_i=k), \quad \forall i \in \{1,\dots,N\}.
\end{align}
To sample from a probability distribution efficiently we implemented a GPU kernel in \\ \fct{sample\_by\_probability} based on
a C algorithm~\citep{smith2002sample}. 
To summarize all $k$ of a point we wrote the  \fct{dcolwise\_dot\_all\_kernel} kernel.
These kernels are working in parallel on all components. Each component is working in a different GPU stream while in each stream the points are aggregated to blocks that are running in parallel. We set a number of 512 threads per block. This parameter can be fine-tuned to optimize the performance based on the GPU hardware being used.

\item Sample sub-cluster assignments for each point: 
\begin{align} \label{eq:sample_sub_labels}
\bar z_i \appropto \sum\nolimits_{h\in\{l,r\}} \pi_{z_i h} f_{\bx}(\bx_i;\bar\theta_{z_i h}) \mathbbm{1}(\bar z_i=h), \quad \forall i \in \{1,\dots,N\}.
\end{align}
\end{enumerate}
$\bar z_i \in \{l,r\}$ variables are (conceptually thought of as ``left'' and ``right'') indicate which of the sub-cluster the $i^th$ point is assigned to.
We have a thin version of the cluster and sub-cluster parameters \code{thin\_cluster\_params}. For a single machine this structure is unneeded since the data already exists in \class{local\_cluster}. However, we maintain it not only for consistency with the \proglang{Julia} package but also 
because this option may be valuable in the future to scale our \proglang{CUDA/C++} implementation to multiple machines.

For efficiency we are not copying the data; rather, we use \code{std::shared\_ptr} between the structures. We have also chunks of the data per GPU in the device memory as part of the \class{gpuCapability} structure. The structure contains the following 3 properties: List of points $\bx_i$ (\code{d\_points}), chunks of the cluster assignments $z_i$ (\code{d\_labels}) and sub-cluster assignments $\bar z_i$ (\code{d\_sub\_labels}).
\begin{Code}
struct gpuCapability
{
	int* d_labels;
	int* d_sub_labels;
	double* d_points;
};

class cudaKernel
{
protected:
	std::map<int, gpuCapability> gpuCapabilities;
};
\end{Code}

\item Propose and Accept Splits:\\
By sampling the sub-clusters, one is able to propose and accept meaningful splits that divide a cluster into its 2 sub-clusters.
\begin{enumerate}[(a)]
\item Propose to split cluster $k$ into its 2 sub-clusters for all $k\in \{1,2,\dots, K\}$.
\item Calculate the Hastings ratio $H$ and accept the split with probability min$(1,H)$:
\begin{align} \label{eq:sample_splits}
H_{\text{split}} = \frac{\alpha \Gamma(N_{kl}) 
f_{\bx}(\bx_{\mathcal{I}_{kl}};\lambda) \cdot \Gamma(N_{kr}) 
f_{\bx}(\bx_{\mathcal{I}_{kr}};\lambda)}{\Gamma(N_k) f_{\bx}(\bx_{\mathcal{I}_k};\lambda)} 
\end{align}
The proposal is done in the function \fct{should\_split\_local} and the split itself is done in the kernel \fct{split\_cluster\_local\_worker}.
\end{enumerate}

\item Propose and Accept Merges:\\
Merges are proposed by merging 2 sub-clusters into one with each of the original clusters becoming a sub-cluster of the new merged cluster.
\begin{enumerate}[(a)]
\item Propose to merge clusters $k_1, k_2$ for all pairs $k_1, k_2 \in \{1,2,\dots, K\}$.
\item Calculate the Hastings ratio $H_{\text{merge}}$,
\begin{align} \label{eq:sample_merges}
H_{\text{merge}} = \frac{\Gamma(N_{k_1}+N_{k_2})}{\alpha\Gamma(N_{k_1}) 
\Gamma(N_{k_2})}
\frac{p(\bx|\hat{z})}{p(\bx|z)}
\times
\frac{\Gamma(\alpha)}{\Gamma(\alpha+N_{k_1}+N_{k_2})} 
\times
\frac{\Gamma(\frac{\alpha}{2}+N_{k_1})\Gamma(\frac{\alpha}{2}+N_{k_2})}{
\Gamma(\frac{\alpha}{2})\Gamma(\frac{\alpha}{2})}\, ,
\end{align}
 and accept the merge with probability min$(1,H_{\text{merge}})$.
The proposal is done in function \fct{should\_merge} and the merge itself is done in the kernel \fct{merge\_clusters\_worker}.

\end{enumerate}

\end{itemize}

\subsection{Optimizing Processing}
\textbf{Two kernels for optimization}\\
In order to optimize the performance of the matrix multiplication which is required in our algorithm we used two different \proglang{CUDA} kernels and decide which one to use based on the size of the $d\times N$ matrix 
(where $d$ is the dimension of the data and $N$ is the number of points). We added a built-in capability to automatically select, at run-time, the best kernel automatically based on $d$, $N$ and the GPU capabilities. For more optimization in case that the best kernel is known we provide an option to set the kernel which can save some time at the beginning of the run (especially with a high $N$ and a high $d$). We measured the optimal kernel on an NVidia Quadro RTX 4000 card. On matrices whose size was below 640,000 size ($d\times N$), Kernel \#1 is optimized and above this number Kernel \#2 is optimized. Kernel \#1 was written in a native way for \proglang{CUDA} to multiply two matrices. Kernel \#2 was written with cublas API which is more optimal for big matrices. 

\textbf{Data Structure}\\
To optimize the operations that needed to be done on the matrices and in the device, all the data in the GPU and in Eigen's structure are stored in a column-major order. In general, in places which we needed to perform operations on all data points, we iterate on the dimensions and run (in parallel) each point calculation on a different GPU core.

\subsection{Parallelism}
The sampler from~\cite{Chang:NIPS:2013:ParallelSamplerDP} was designed with massive parallelization in mind. Thus, most of its operations are parallelizable. Particularly, sampling cluster parameters is parallelizable over the clusters, sampling assignments can be computed independently for each point, and cluster splits can be proposed in parallel. Thus, a multi-core implementation is quite straightforward, with the caveat that one needs to be careful with merges; \ie, to prevent more than 2 clusters merge into the same single cluster; \eg, if clusters 1 and 2 are merged at the same time when clusters 2 and 3 are merged, this would imply the three clusters (1,2, and 3) would be merged into a single one. However, this would be inconsistent with the underlying model. 
Recall that \cite{Chang:NIPS:2013:ParallelSamplerDP} released a multi-core single-machine \proglang{C++} implementation (for CPU). 
The nature of the algorithm, however,  let us build an implementation that goes beyond their original implementation, by allowing parallelization across multiple machines, not just multiple CPU cores. In turn, this enables us to not only leverage more computing power and gain speedups but also provide practical benefits in terms of memory and storage. For example, our Julia implementation can be used within a distributed network of weak agents (\eg, small robots collecting data). It also never transfers data (which is expensive and slow); rather, we transfer only sufficient statistics and parameters. Thus, the proposed implementation is also well suited for a network of low-bandwidth communication. Likewise, on a single machine,  we are able to use the GPU's cores powers and Nvidia streams and asynchronous calls.

\subsubsection{Multiple GPU Streams}
\begin{figure}[t!]
\centering
\includegraphics[width=\linewidth]{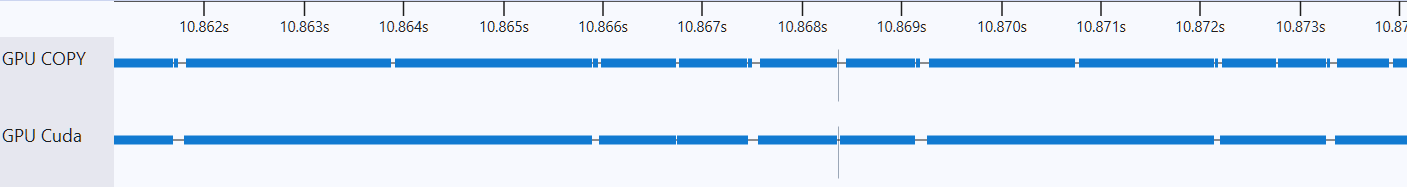}
\caption{\label{fig:cuda-multi-streams} \proglang{CUDA} running in multi streams. Copy and Kernels are working in parallel. Blue color indicates
an active kernel. The horizontal axis stands for time (in [sec]).}
\end{figure}

Most of the code was written as separated individual streams. Each stream contains sequences of operations that execute in issue-order on the GPU. \proglang{CUDA} operations in different streams may run concurrently. In many places we tied the stream to a specific cluster. Thus, we could run close to $O(1)$ instead of $O(K)$. We took advantage of \proglang{CUDA}'s asynchronous memory allocation API which was exposed lately in version 11. We were able to allocate and deallocate the parameters as part of the sequence of operations and in this way to improve the concurrency of other kernels that were running on other streams. In~\autoref{fig:cuda-multi-streams} we show an example of memory allocation operations that were performed on the GPU in parallel and at the same time that the kernel operation was running.

\subsubsection{Multiple GPU Cards}
We tested the performance of our implementation by parallelizing the processing across multiple GPUs. The logic that we adopt was to parallelize the parts that we were running with multi-streams on multi-GPUs. We implemented a container with all available GPUs: \code{std::map<int, gpuCapability> gpuCapabilities}. In the places that we sample the cluster and sub-cluster parameters and where the sufficient statistics are been calculated, instead of creating a stream on the same GPU we took the next available GPU by calling to the function \code{pick\_any\_device} and created a stream on the current selected device. Our hypothesis was that in places which $K$ is large there should be more impact. Our observation when we tested 2 GPUs (of type Quadro RTX 4000) was that the performance is not better. Consequently, we disabled that option by default in the \code{cudaKernel::init} function using the following line: \code{numGPU = 1}. On a different hardware and different GPU types, however, it is possible that this option will yield better results. Thus, the user may want to experiment with code{numGPU > 1}, depending on the available hardware. 

\subsection{Runtime Complexity}
We now go throughout the \proglang{CUDA/C++} runtime complexity step by step based on the algorithm. The symbol $G$ below denotes the number of GPU cores. \\
{\bf For each iteration of a restricted Gibbs sampling:}
Sampling the cluster and sub-cluster parameters (including weights) takes constant time 
for each cluster. In \proglang{Julia} it is parallelized over $P$ processes on the master. This takes $O(K \times d/P)$ time. In \proglang{CUDA/C++} we observed that using \code{'omp parallel for'} (for $K$) when sampling the cluster parameters is slower than performing it sequentially. The root cause is the fact that Eigen (which we rely on for matrix manipulation) already uses multiple CPU cores and it also uses the GPU efficiently in each one of its threads. Thus, running over the $K$ clusters in parallel is slower than sequentially. Therefore the complexity of \proglang{CUDA/C++} is $O(K \times d)$. Sampling the cluster weights is also done in constant time.
Copying cluster and sub-cluster weights and parameters from host to device is parallelized over \proglang{CUDA} streams over $K$ clusters.
The process is parallelized over $K$ clusters and over $G$ GPU cores so it takes $O(K \times d/G)$.
Sampling the cluster assignments takes $O(N \times K \times T/G)$ time, where $T$ depends on the observation model, \eg for multivariate Gaussian observation model with NIW prior $T=d^2$, wherein for a multinomial observation model with Dirichlet prior $T=d$.
Sampling the sub-cluster assignments takes $O(N \times T/G)$.
Updating the cluster and sub-cluster sufficient statistics can be split up into 3 steps. The first step is for all streams to calculate the sufficient statistics for the data it is in charge of which is common to all kinds of distributions. This step takes $O(N/G)$ time. 
The second step is to calculate the sufficient statistics which is unique per the distribution. It is done in parallel to each sub-cluster $(l,r)$ and to the cluster. This step takes $O(N \times T/G)$ time. The third step is to aggregate across all streams. This step takes $O(K \times d)$ time.
Overall, this takes $O(N \times K \times T/G)$ + $O(N \times T/G) + O(K)$ time. So total: $O(N \times K \times T/G)$.

{\bf Splits:}
Proposing splits by looking at each cluster is $O(K)$, noting that we can drop the $T$ here as we use previously calculated values.
Processing all the accepted splits requires updating the sufficient statistics which could take at the worst case $O(N/G) + O(K)$ if all clusters are split.

{\bf Merges:}
Proposing merges by inspecting each cluster pair is $O(K^2)$, as in the splits, we can drop the $T$, by using the previously calculated values. Processing all the accepted merges also requires updating sufficient statistics. The worst case (\ie, if all clusters are merged) is thus $O(N) + O(K)$.

To summarize, the total runtime complexity is $O(K) + O(N \times K \times T/G) = O(N \times K \times T/G)$

\subsection{Memory Complexity}
We now look at the amount of memory used on the GPU. The data is stored as one array, so we have $O(d \times N)$ on the GPU. The amount of data for the labels and sub labels is $O(N)$. Each stream also has a copy of the cluster and sub-cluster parameters which takes $O(K)$ space. We also have to aggregate sufficient statistics for each cluster after sampling the assignments, taking $O(K \times T)$. Hence, we have a total memory usage of $O(d \times N)$. Since usually $N \gg K$, the memory overhead is insignificant in comparison to the data itself.

\section{Illustrations}

\subsection{Synthetic Data: Dirichlet Process Gaussian Mixture Model (DPGMM)}
We tested our implementation on small and large synthetic GMM datasets. We wrote a class \code{data\_generators} to generate the synthetic datasets. We ran 112 tests with different parameters: $N$ ($10^3$,$10^4$,$10^5$,$10^6$), $d$ (2,4,8,16,32,64,128) and $K$ (4,8,16,32). For each test we ran 100 iterations (sufficed for convergence
in all methods). We repeated each test 10 times with the same random Gaussian synthetic data that we generated and then averaged the resulting NMI scores and running times. In each test we compared \proglang{CUDA/C++}, \proglang{Julia}, and a Bayesian Gaussian Mixture from sklearn. The latter, like our implementations, infers $K$ with the rest of the model. However, unlike our implementations, this sklearn's model requires an upper bound on $K$.

During our testing we observed that, on a single machine, our \proglang{CUDA/C++} implementation is faster than our \proglang{Julia} 
implementation in most of the cases. The cases in which \proglang{Julia} was faster were when running with a low $N$ (up to 10K) and a low $d$ (up to 32). When the data is larger (\eg, when $N=10^6$ or $d=128$) the \proglang{CUDA/C++} solution is 2.5 times faster than \proglang{Julia} on average. In general, when $N \times d \times K$ is high the \proglang{CUDA/C++} solution is faster than the \proglang{Julia} solution. Both \proglang{Julia} and \proglang{CUDA/C++} are faster on average than sklearn solution:  on average, \proglang{Julia} was 2.6 times faster and the \proglang{CUDA/C++} was 5.3 times faster. Moreover, in a few cases the \proglang{CUDA/C++} was 35 times faster than sklearn.

\begin{figure}[t!]
\centering
\begin{tikzpicture}
  \node (img1)  {\includegraphics[width=\linewidth]{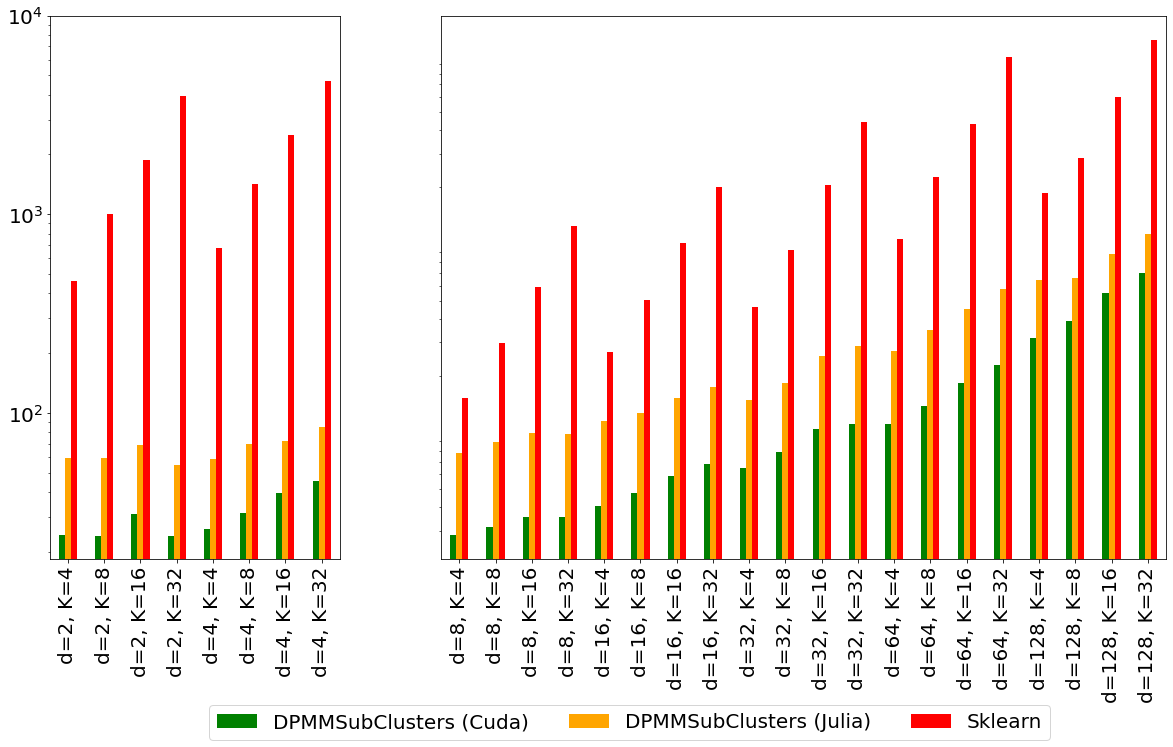}};
  \node[left=of img1, node distance=0cm, rotate=90, anchor=center,yshift=-0.8cm, xshift=1cm] {Time (logarithmic scale) [sec]};
 \end{tikzpicture}
\caption{\label{fig:DPGMM-time-1million} DPGMM synthetic data, time, $N=10^6$. 
In the right panel, due to sklearn's slowness, we had to give it
the unfair advantage of using the true $K$ as the upper bound of the number of clusters.  
}
\end{figure}

\begin{figure}[t!]
\centering
\begin{tikzpicture}
  \node (img2)  {\includegraphics[width=\linewidth]{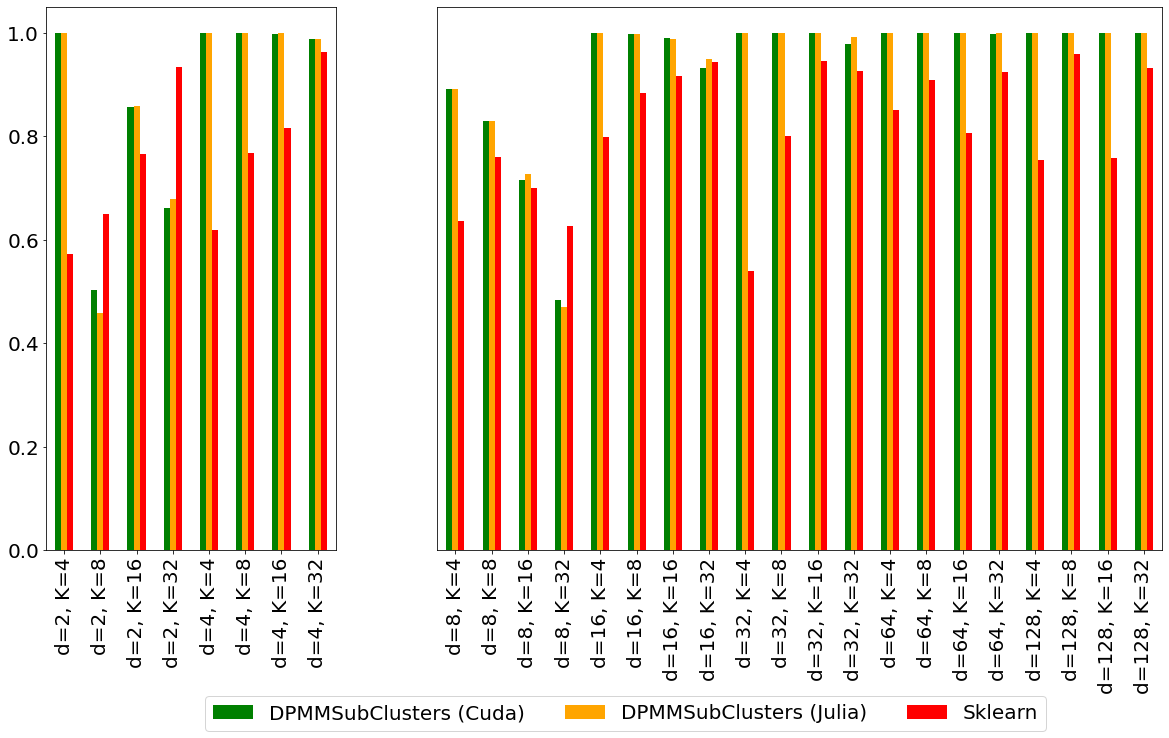}};
  \node[left=of img2, node distance=0cm, rotate=90, anchor=center,yshift=-0.8cm, xshift=0.5cm] {NMI Score};
 \end{tikzpicture}
\caption{\label{fig:DPGMM-NMI-1million} DPGMM synthetic data, NMI, $N=10^6$. In the right panel, due to sklearn's slowness, we had to give it
the unfair advantage of using the true $K$ as the upper bound of the number of clusters.}
\end{figure}

In \autoref{fig:DPGMM-time-1million} we demonstrated the time comparison for $N=10^6$. In the left side when $d\le 4$ we see that our solutions were faster than sklearn. Due to the slowness of sklearn for the cases when $d>4$ we set in the right side of the figure an easier target for sklearn and set sklearn's upper bound for $K$ to the true $K$ (without doing it, sklearn's running time for a high $d$ was impracitically slow,
many orders of mangitude slower than our implementations). 
It is obvious from the figure than even when we gave a big advantage to sklearn, our implementations still beat it. For accuracy we measured the NMI for each case. NMI comparison is displayed in \autoref{fig:DPGMM-NMI-1million}. Again we split the test into two parts. 
The left side of the figure describes a fair comparison where
all the methods were given the same conditions. 
The right side describes the case sklearn enjoyed the advantage
of a reduced complexity achieved by using the true $K$
as the upper bound. Despite that advantage, our solutions still yielded better results almost in all cases. All tests can be reproduced by running the dpmmwrapper.ipynb Jupiter Notebook from \url{https://github.com/BGU-CS-VIL/dpmmpython}.
Because of the slowness of sklearn, the notebook as is might take many days. You can consider to reduce the number of permutations that we tested for $N$, $d$ and $K$ or to reduce the number of repeats.

\subsection{Synthetic Data: Dirichlet Process Multinomial Mixture Model (DPMNMM)}
\begin{figure}[t!]
\centering
\begin{tikzpicture}
  \node (img3)  {\includegraphics[width=\linewidth]{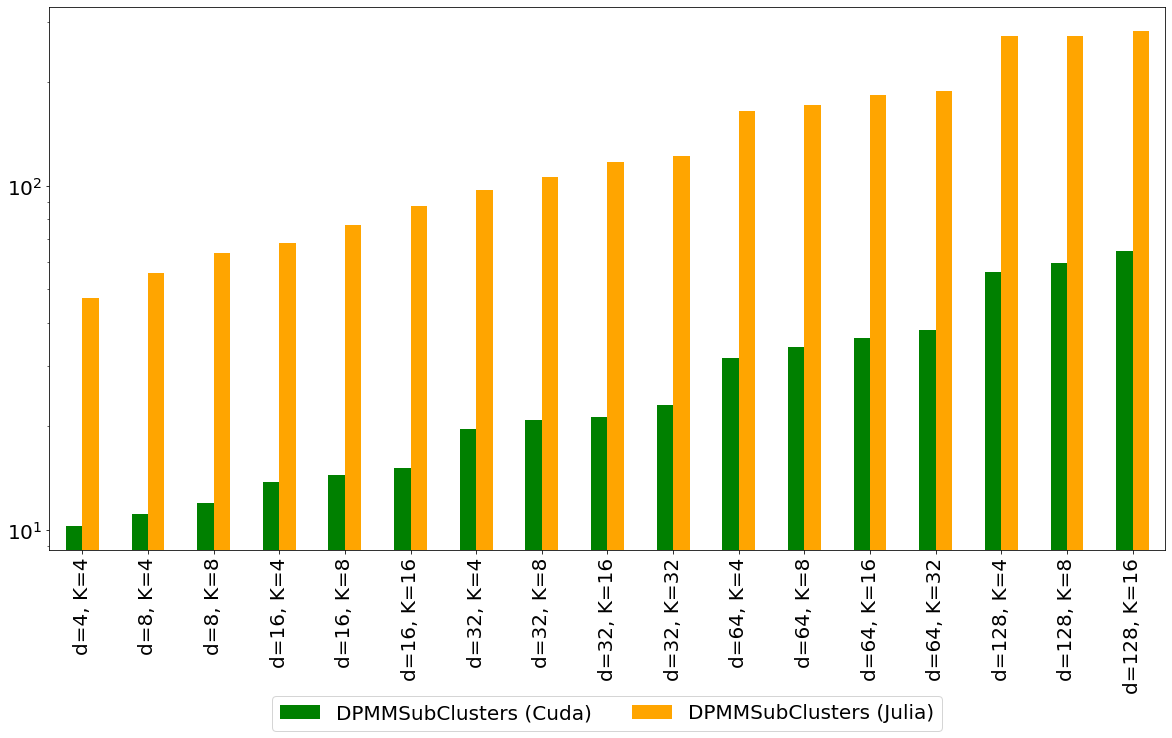}};
  \node[left=of img3, node distance=0cm, rotate=90, anchor=center,yshift=-0.8cm, xshift=1cm] {Time (logarithmic scale) [sec]};
 \end{tikzpicture}
\caption{\label{fig:DPMNMM-time-1million} DPMNMM synthetic data, time, $N=10^6$. 
}
\end{figure}

\begin{figure}[t!]
\centering
\begin{tikzpicture}
  \node (img4)  {\includegraphics[width=\linewidth]{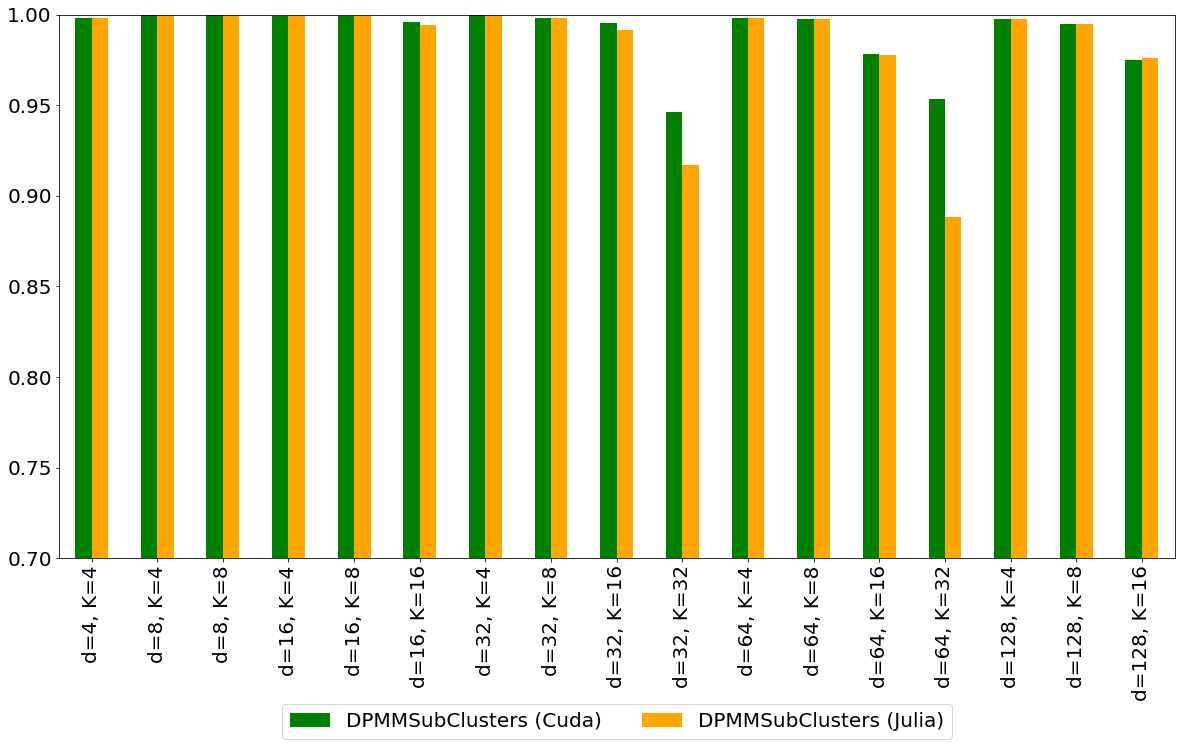}};
  \node[left=of img4, node distance=0cm, rotate=90, anchor=center,yshift=-0.8cm, xshift=1cm] {NMI Score};
 \end{tikzpicture}
\caption{\label{fig:DPMNMM-NMI-1million} DPMNMM synthetic data, NMI, $N=10^6$.}
\end{figure}

We ran 72 tests with different parameters: $N$ ($10^3$,$10^4$,$10^5$,$10^6$), $d$ (4,8,16,32,64,128) and $K$ (4,8,16,32) while $d\geq K$. For each test we ran 100 iterations (sufficed for convergence). We repeated 10 times each test with the same random multinomial synthetic data and then averaged the
NMI results and running times. In each test we compared
our \proglang{CUDA/C++} and \proglang{Julia} solutions to each other (sklearn does not support
multinomial components when the number of components is unknown). 
Unlike in the case of Gaussian components, 
here, with multinomials, we observed that our \proglang{CUDA/C++} solution was uniformly faster than our \proglang{Julia} solution. On average the former was 5 times faster than the latter. The results (using $N=10^6$) are displayed in~\autoref{fig:DPMNMM-time-1million}. The corresponding NMI scores apear in~\autoref{fig:DPMNMM-NMI-1million}.

\subsection{Real Data}
\begin{figure}[t!]
\centering
\begin{tikzpicture}
  \node (img5)  {\includegraphics[width=150mm, height=80mm]{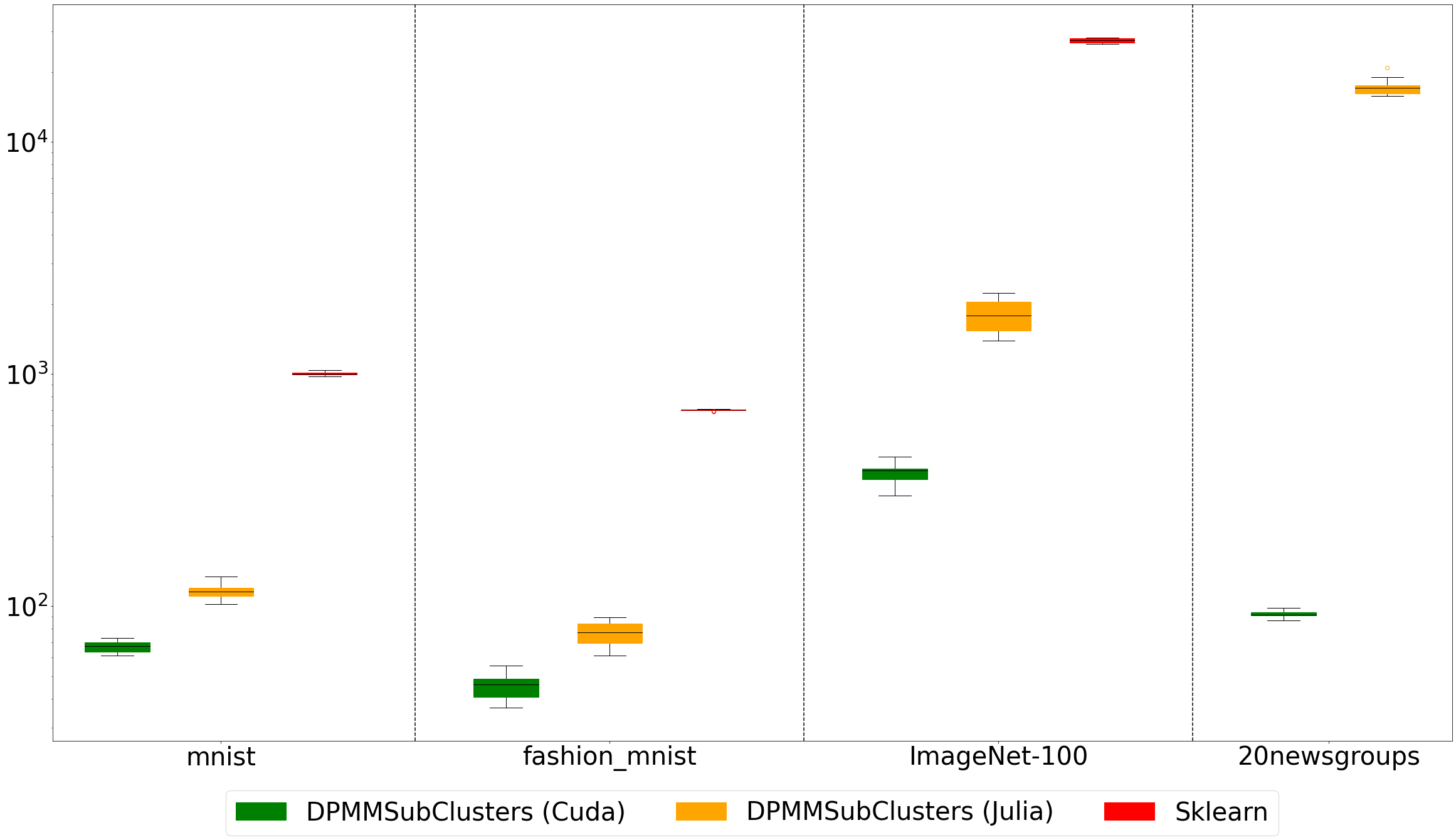}};
  \node[left=of img5, node distance=0cm, rotate=90, anchor=center,yshift=-0.8cm] {Time (logarithmic scale) [sec]};
 \end{tikzpicture}
\caption{\label{fig:DP-realdata-time} DPMM on real data: running times. In the 20newsgroup dataset (where the data dtype is discrete) the  components are Multinomials. In the other datasets the components are Gaussians.}
\end{figure}

\begin{figure}[t!]
\centering
\begin{tikzpicture}
  \node (img6)  {\includegraphics[width=150mm, height=80mm]{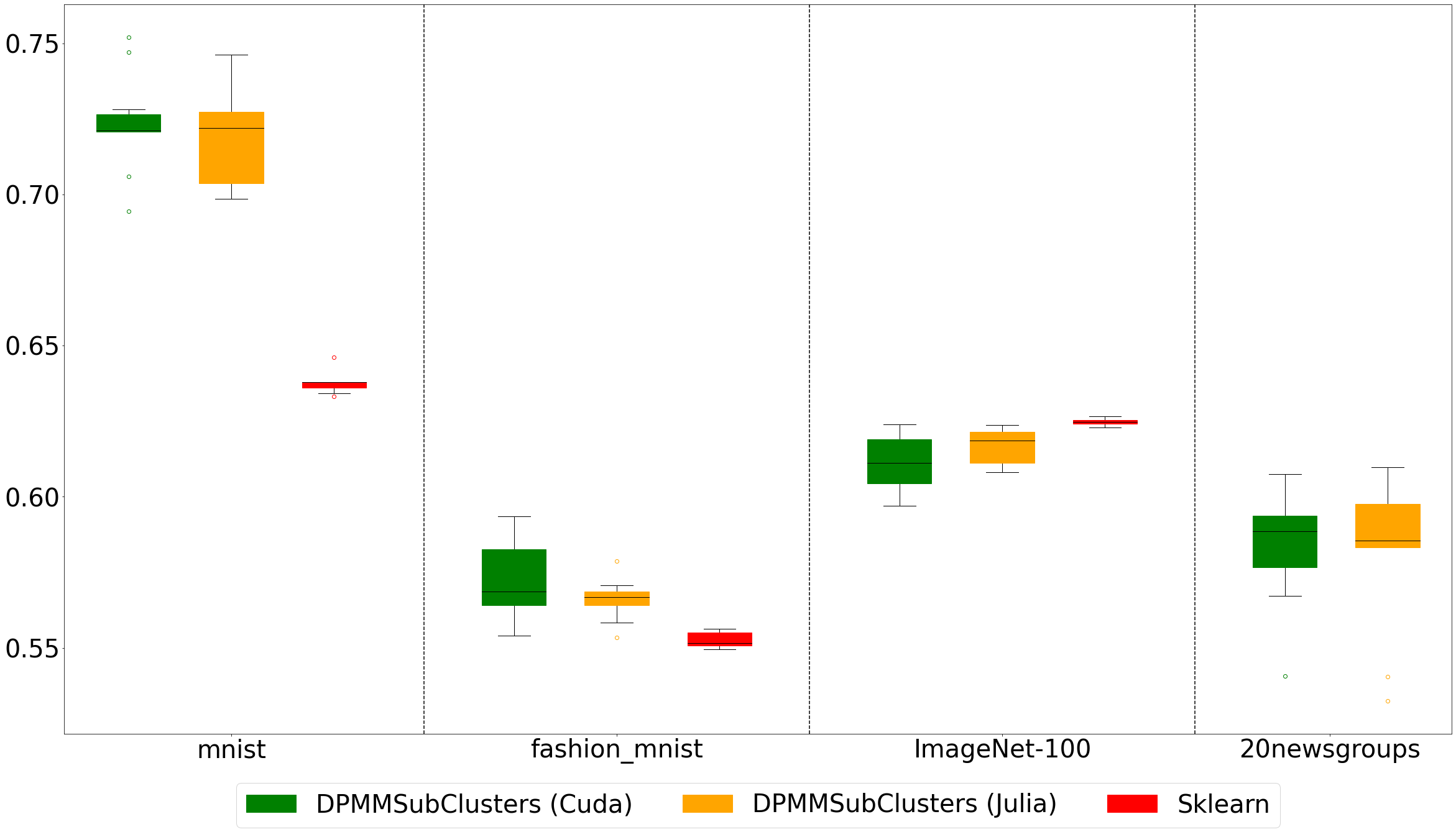}};
  \node[left=of img6, node distance=0cm, rotate=90, anchor=center,yshift=-0.8cm] {NMI Score};
 \end{tikzpicture}
\caption{\label{fig:DP-realdata-NMI} DPMM on real data: NMI scores. In the 20newsgroup dataset (where the data dtype is discrete) the  components are Multinomials. In the other datasets the components are Gaussians.}
\end{figure}

We also tested our solution on real datasets, where, as pre-processing,
we used Principal Component Analysis (PCA) to reduce the dimensionlity. For DPGMM we used the following datsets: \textbf{mnist} ($N=60000$,$d=32$,$K=10$); \textbf{fashion mnist} ($N=60000,d=32,K=10$) and \textbf{ImageNet-100} ($N=125000$, $d=64$, $K=100$). We compared our \proglang{CUDA/C++} and \proglang{Julia} implementations with sklearn BayesianGaussianMixture. For DPMNMM we used \textbf{20newsgroups} ($N=11314,d=20000,K=20$), and compared our \proglang{CUDA/C++} and \proglang{Julia} implementations to each other (sklearn does not support
multinomial components when the number of components is unknown). 
In the DPMNMM case, when the dimension was very high ($d=20,000$)
the \proglang{CUDA/C++} package was 188 times faster than \proglang{Julia}. In~\autoref{fig:DP-realdata-time} we compared the running times on each dataset between the different packages. It is clear from the figure that our \proglang{CUDA/C++} package is much faster from the two others, and that our Julia package is much faster than sklearn.  In~\autoref{fig:DP-realdata-NMI} we displayed the NMI comparison the different packages. On \textbf{ImageNet-100}, our \proglang{CUDA/C++} and \proglang{Julia} packages are almost equal sklearn (with a minor difference of 0.013). In all other cases, 
our solutions were more accurate than sklearn. Morover, even though, 
on \textbf{ImageNet-100}, our solutions did not score higher NMI values than  sklearn, the $K$ value that we predicated was much more accurate. 
That is, while the true $K$ was 100, sklearn predicated $K=500$ (which was the value of the upper bound it was given). In contrast we predicted on average $K=96.8$ (with a standard deviation of 17.78).

\section{Summary and Discussion}

We extended the DPMM inference method from~\cite{Chang:NIPS:2013:ParallelSamplerDP} by efficient and easily-modifiable implementations for high-performance distributed sampling-based inference. Our software can be adopted by practitioners in an easy way where the user is free to choose between either a multiple-machine, multiple-core, CPU implementation (written in \proglang{Julia}) and a multiple-stream GPU implementation (written in \proglang{CUDA/C++}). We also provide an optional \proglang{Python} wrapper which hides the CPU and GPU implementations as a single package with one interface. The proposed implementation supports both Gaussian and Multinomial components. However, 
it is also  easy to extent the code to support other exponential families. We also tested the proposed implementation for learning models from real datasets as mnist, fashion mnist, ImageNet-100 and 20newgroups. We compared our solution to sklrean and showed that for large and high dimensions our solution achieves the same, or better, accuracy while being much faster. Empirically, we found that when using high-dimensional Gaussians on a single machine our GPU package is faster than any other solution that is currently publicly available. For Multinomial components, our GPU package showed even better results which are by at least two order of magnitude faster than any other existing solution. We provided a solution for cross operating systems platforms (Windows \& Linux). Our tests can be easily reproduced via a Jupiter Notebook included with our \proglang{Python} package.

\section{Computational details}

We ran our tests on two different hardware. One with strong CPU, more RAM and GPU from GTX family: i7-6800K CPU, 3.40GHz with 12 cores, 64GB RAM, GeForce GTX 1070. Second configuration with slower CPU, less RAM and GPU from RTX family: E5-2620 v3 CPU, 2.40GHz with 12 cores, 32GB RAM, Quadro RTX 4000. In both cases we used the same code. We used \proglang{CUDA} driver version 11. We tested on Windows 10 and Linux Ubuntu 18.04 and 21.04. The \proglang{Python} version that we used for the wrapper was 3.8. In the \proglang{CUDA/C++} package for windows we used \pkg{OpenCV} version 4.5.2 to demonstrate visualization in 2D of the clustering process iteration by iteration.

\section{Open Sources}

\begin{table}[ht]
\centering
\begin{tabular}{ p{3.5cm} p{5.5cm} p{5.5cm} } 
\hline
Open Source             & Usage                                     & Link \\ \hline
cnpy	                & Read and write models from npy format     & \url{https://github.com/rogersce/cnpy} \\
zlib		            & Required by cnpy                          & \url{https://github.com/madler/zlib} \\
dirichlet\_distribution & Dirichlet distribution                    & \url{https://github.com/gcant/dirichlet-cpp} \\
logdet                  & Log determinant for Eigen using Cholesky  & \url{https://gist.github.com/redpony/fc8a0db6b20f7b1a3f23} \\
vcflib                  & Log(gamma) and multinormal sampling       & \url{https://github.com/vcflib/vcflib} \\
eigen                   & Matrix and vector operations              & \url{https://eigen.tuxfamily.org} \\
stats                   & Sampling from an inverse-Wishart          & \url{https://www.kthohr.com/statslib.html} \\
gcem                    & Required by stats                         & \url{https://github.com/kthohr/gcem} \\
jsoncpp                 & Read and write Json files	                & \url{https://github.com/open-source-parsers/jsoncpp} \\
MIToolbox               & Calculates NMI                            & \url{https://github.com/Craigacp/MIToolbox} \\
opencv	                & Drawing (for debugging)                   & \url{https://opencv.org/} \\ \hline

\end{tabular}
\caption{\label{tab:cuda/c++_opensource} \pkg{C++} open-source packages used in the proposed implementation}
\end{table}

\begin{table}[ht]
\centering
\begin{tabular}{ p{3.5cm} p{5.5cm} p{5.5cm} } 
\hline
Open Source             & Usage                                     & Link \\ \hline
Clustering.jl           & Evaluation metrics                        & \url{https://github.com/JuliaStats/Clustering.jl} \\
DistributedArrays.jl    & Distribute  Computations                  & \url{https://github.com/JuliaParallel/DistributedArrays.jl} \\
Distributions.jl        & Probability Distributions                 & \url{https://github.com/JuliaStats/Distributions.jl} \\
JLD2.jl                 & Saving and restoring checkpoints          & \url{https://github.com/JuliaIO/JLD2.jl} \\
NPZ.jl                  & Compatability with Numpy data             & \url{https://github.com/fhs/NPZ.jl} \\
SpecialFunctions.jl     & Log Gamma function                        & \url{https://github.com/JuliaMath/SpecialFunctions.jl} \\
\hline
\end{tabular}
\caption{\label{tab:julia_opensource} \pkg{Julia} open-source packages used in the proposed implementation}
\end{table}

In \autoref{tab:cuda/c++_opensource} we specify all the open sources that we use in the \pkg{CUDA/C++} package.
In \autoref{tab:julia_opensource} we specify all the open sources that we use in the \pkg{Julia} package.

\clearpage
\bibliography{refs}

\newpage


\end{document}